\documentclass[10pt]{article}
\usepackage{arxiv}
\usepackage[authoryear,round]{natbib}

\usepackage{xcolor}

\usepackage{amsmath,amsfonts,bm}

\def\eqref#1{equation~\ref{#1}}

\def\1{\bm{1}}

\DeclareMathAlphabet{\mathsfit}{\encodingdefault}{\sfdefault}{m}{sl}
\SetMathAlphabet{\mathsfit}{bold}{\encodingdefault}{\sfdefault}{bx}{n}

\renewcommand{\eqref}[1]{\textup{(\ref{#1})}}
\usepackage[T1]{fontenc}
\usepackage{url}
\usepackage{graphicx}
\usepackage{booktabs}
\usepackage{array}
\usepackage{tabularx}
\usepackage{wrapfig}
\usepackage{algorithm}
\usepackage{algpseudocode}
\usepackage[hypertexnames=false,hidelinks]{hyperref}
\hypersetup{
  pdftitle={Direction-Aware Token Credit Assignment with Reward-Model Gradients},
  pdfauthor={Yikuan Hu, Jifeng Zhu, Minle Su, Jiaqin Yue, Gujie Shao, Yanzhe Chen, Haoxuan Jia, Haoran Xu, Wangzhou Dai},
  pdfsubject={Preprint},
  pdfkeywords={reinforcement learning, token credit assignment, reward model, GRPO}
}
\graphicspath{{resource/}}
\title{Direction-Aware Token Credit Assignment with Reward-Model Gradients}
\author{%
\makebox[\dimexpr\textwidth-2\tabcolsep\relax][c]{%
\begin{tabular}[t]{@{}c@{}}
\textbf{Yikuan~Hu$^{1\diamondsuit}$ \quad Jifeng~Zhu$^{2\diamondsuit}$ \quad Minle~Su$^{1\diamondsuit}$ \quad Jiaqin~Yue$^{3}$}\\
\textbf{Gujie~Shao$^{2}$ \quad Yanzhe~Chen$^{2}$ \quad Haoxuan~Jia$^{4}$ \quad Haoran~Xu$^{3}$ \quad Wangzhou~Dai$^{1\dagger}$}\\[0.25em]
\normalfont $^{1}$Nanjing University \quad $^{2}$National University of Singapore \quad $^{3}$Tsinghua University\\
\normalfont $^{4}$Nanyang Technological University
\end{tabular}%
}
}

\date{}
\makeatletter
\renewcommand{\@maketitle}{%
  \vbox{%
    \hsize\textwidth
    \linewidth\hsize
    \vskip 0.2in
    {\LARGE\scshape\raggedright\@title\par}
    \vskip 0.25in
    {\centering\normalsize\@author\par}
    \vskip 0.2in
  }%
}
\makeatother
\AtBeginDocument{\newgeometry{margin=1.05in,headheight=14pt,headsep=25pt,footskip=30pt}}
\begin{document}
\pagestyle{fancy}
\maketitle
\thispagestyle{fancy}
\begingroup
\renewcommand{\thefootnote}{\ensuremath{\diamondsuit}}
\begin{NoHyper}
\footnotetext{Equal contribution.}
\renewcommand{\thefootnote}{\ensuremath{\dagger}}
\footnotetext{Corresponding author.}
\end{NoHyper}
\renewcommand{\thefootnote}{}
\footnotetext[0]{Project page: \url{https://datc-project.github.io/}.}
\endgroup
\setcounter{footnote}{0}
\begin{abstract}
GRPO shares a response-level advantage across tokens, leaving unresolved how strongly each token should contribute to learning. Updates induced by the shared advantage reinforce or suppress the sampled token at each position, but these updates do not always point toward increasing reward at the token level. This motivates a \textbf{work-based view of credit}: just as work depends on the component of force along a displacement, token credit should reflect the local reward change along the policy update direction. We introduce \textbf{Direction-Aware Token Credit Assignment (DATC)}, which maps frozen reward-model gradients into policy logit space and allocates weights by evaluating the alignment between the reward direction and GRPO's signed sampled-action update direction, while preserving the sign and mean of the response-level advantage. DATC outperforms GRPO in all four settings across two model families.
\end{abstract}

\section{Introduction}
\label{sec:introduction}

\begin{wrapfigure}{r}{0.4\textwidth}
    \centering
    \includegraphics[width=\linewidth]{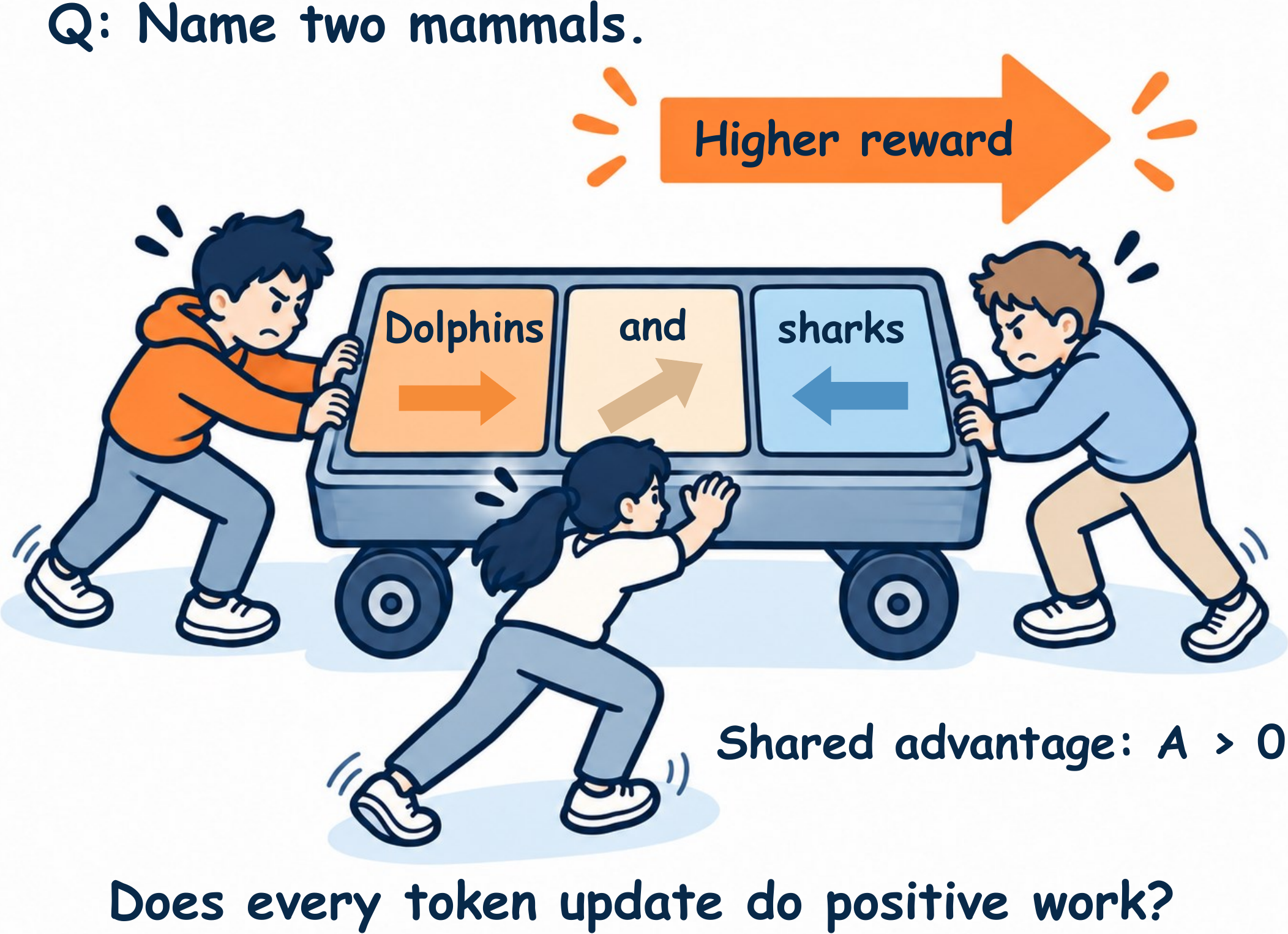}
    \caption{\textbf{A work-based view of token credit.}
    DATC assigns larger weights to token updates with
    higher predicted local reward gains.}
    \label{fig:teaser}
\end{wrapfigure}

Preference alignment of language models relies on evaluating complete responses: determining whether a response is useful or follows instructions often requires contextual judgment. A reward model (RM) converts such judgments into response-level scores, which reinforcement learning uses to improve the generation policy~\citep{zhong2024dpo}. Group Relative Policy Optimization (GRPO)~\citep{shao2024deepseekmath} reinforces better responses and suppresses worse ones based on their relative rewards for the same input. Within a response, all positions share a response-level advantage coefficient, which reinforces or suppresses sampled tokens with the same coefficient. However, a shared response-level advantage does not imply that updates at all token positions are equally beneficial~\citep{ren2026effect,wang2026emergent}. The credit assignment problem studied here is how to allocate learning strength according to the reward utility of each position's update.

Existing studies of token-level credit assignment mainly follow two approaches. One uses reward-model signals to assign token credit: RED~\citep{li2025red} redistributes response-level rewards using differences between successive prefix scores, while P2T~\citep{wu2026unlocking} uses RM input gradients to approximate the effects of token replacement. The other derives the relative utility of token updates from optimization objectives. Examples include VCORE~\citep{gong2026vcore}, which evaluates token utility in SFT through gradient alignment, and DelTA~\citep{zhang2026delta}, which adjusts RL updates using the gradient structure of positive and negative advantages. We further focus on local reward utility along the GRPO update direction under the current policy. To this end, we propose \textbf{Direction-Aware Token Credit Assignment (DATC)}. DATC adopts a \textbf{work-based view of credit}: it allocates learning strength within a response by evaluating the work done by the current token's signed update direction (force) along the RM reward-improvement direction (displacement direction).

To turn this work-based view of credit into a computable token signal, we characterize the reward displacement direction and the force of the policy update in the same logit space. Specifically, our surrogate analysis starts from minimizing the negative RM reward in the end-to-end actor--RM process. On a fixed sampled response, we use a straight-through (ST) surrogate~\citep{bengio2013estimating} to connect discrete tokens to the frozen RM and propagate the standardized reward gradient back to the actor's logit space, obtaining the reward gradient $g_t$. It characterizes the direction and rate of local reward increase and corresponds to the reward-improving update ``displacement'' in the work analogy. Meanwhile, the sampled action's log-probability gradient $k_t$ characterizes the direction that increases its probability. The response-level advantage determines whether to reinforce along this direction or suppress in the opposite direction, thereby specifying the ``force'' of the policy update. We thus consider the following local question: \textbf{when logits are updated along the direction induced by GRPO, how much work does the update's ``force'' do along the reward-improving displacement?} In a local surrogate model with independent logits, a first-order expansion yields directional credit $d_t=\langle g_t,k_t\rangle$. After accounting for the update direction and scale, the predicted standardized reward increment at position $t$ is $(\eta/T)A d_t$. Hence, after removing the common step-size factor, $A d_t$ measures the signed predicted work of this token update: a positive value indicates an increase in the local surrogate reward, whereas a negative value indicates a decrease. DATC allocates training shares within a response accordingly, assigning larger weights to positions with higher signed predicted work. KL regularization controls the allocation's deviation from original GRPO, yielding the final token-level advantage:\footnote{$A$ is the response-level advantage, and $T\,\operatorname{softmax}_t(Ad/\tau)$ gives the direction-aware token weights. $T$ is the number of valid response tokens, and $\tau$ is the allocation temperature for that response.}

\[
\boxed{
\widetilde A_t
= A\cdot T\,\operatorname{softmax}_t\!\left(Ad/\tau\right)
}.
\]

Furthermore, we prove that the end-to-end surrogate gradient has an advantage-weighted structure consistent with REINFORCE, and that directional credit $d_t$ aggregates its local reward advantages along the sampled-action direction of the current policy. These advantages describe each candidate action's local reward difference relative to the current policy's average. They contribute to the position's credit according to the directional relationship between the candidate action and the sampled action. Figure~\ref{fig:workflow} illustrates the overall DATC workflow. Our main contributions are as follows:

\begin{figure}[t]
\centering
\includegraphics[width=\linewidth]{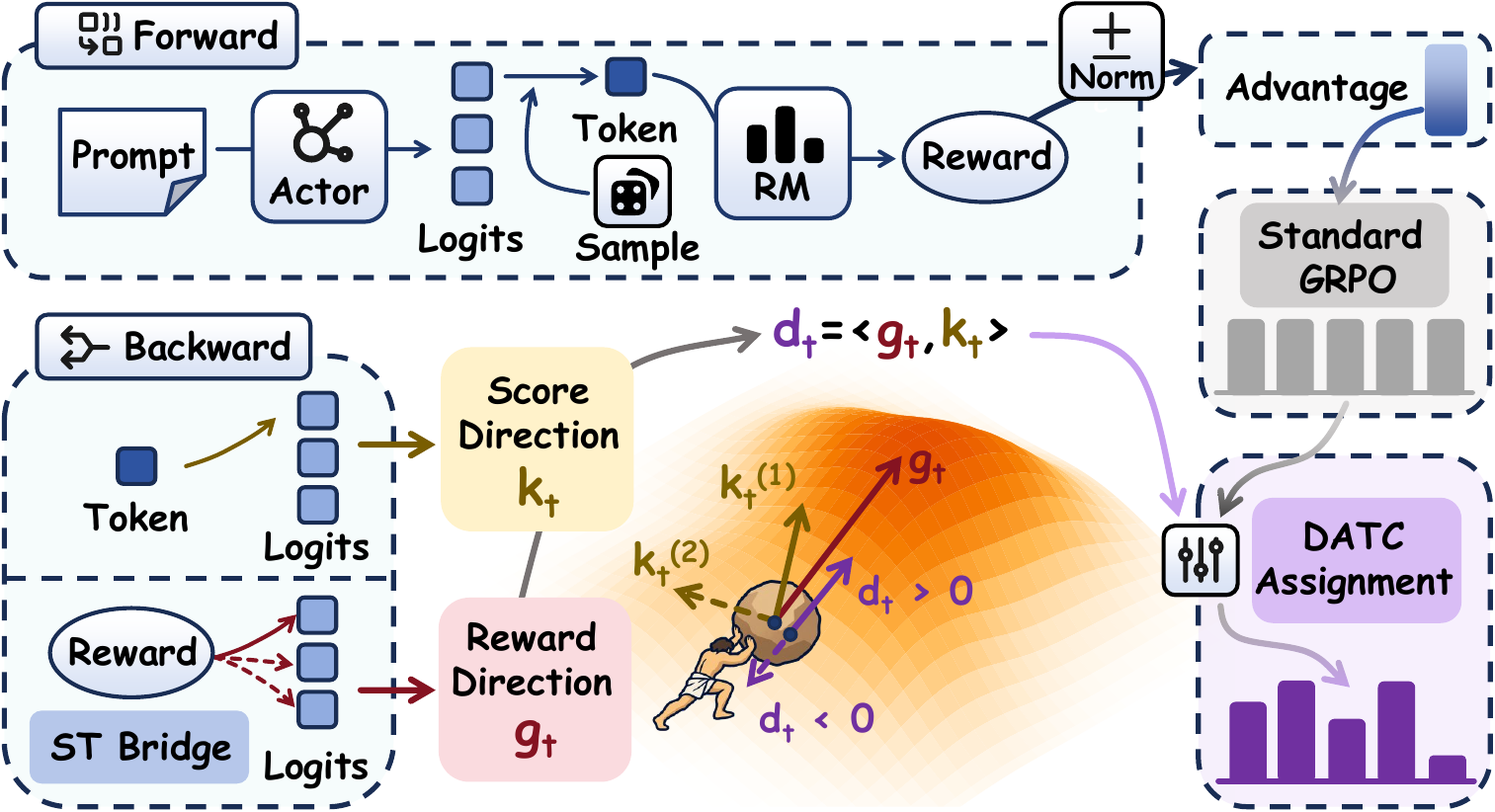}
\caption{Overview of DATC. DATC combines the reward-improvement direction with the GRPO update direction to allocate token-level advantages, assigning larger weights to updates with higher predicted local reward gains.}
\label{fig:workflow}
\end{figure}

\begin{itemize}
\item We propose Direction-Aware Token Credit Assignment, a GRPO token credit assignment method based on a work-based view of credit. It allocates token-level advantages by combining the local reward direction with the sampled-action direction of the current policy.
\item We prove that the end-to-end surrogate gradient has an advantage-weighted structure consistent with REINFORCE~\citep{williams1992simple}, and that DATC's directional credit aggregates advantages defined by the RM's local linear reward model along the sampled-action direction of the current policy.
\item Across Base and Instruct settings of the Qwen3-14B and Llama-3.1-8B model families, DATC achieves higher mean scores than GRPO on all four evaluation metrics. Ablations further support the key roles of directional information and the correspondence between weights and positions in these performance gains.
\end{itemize}

\section{Related Work}
\label{sec:related-work}

\textbf{Fine-grained rewards and value estimation.} Prior work obtains fine-grained feedback through process supervision~\citep{lightman2024let}, implicit process rewards~\citep{yuan2024free,cui2025process}, and token-level reward learning for policy optimization~\citep{chen2025discriminative,zhong2024dpo}. Other methods estimate conditional values and gains in success probability from sampled continuations~\citep{kazemnejad2024vineppo,guo2026segment,setlur2025rewarding}. Building on these fine-grained evaluations, we further study how to allocate group-relative advantages according to the local reward effects of token updates.

\textbf{Token credit from model-internal signals.} Policy entropy, attention, and differences in prefix scores have been used for token selection and reward shaping~\citep{chan2024dense,li2025red,tan2025gtpo}. Gradient attribution from reward or judge models and final-answer sensitivity also inform token credit and advantage reshaping~\citep{wu2026unlocking,zhang2026grad2reward,li2026outcome,pala2026grail}. Policy updates redistribute probability among competing candidate tokens. Evaluating the reward effect of reinforcing or suppressing a sampled token therefore requires connecting input importance to this probability adjustment.

\textbf{Token reweighting from optimization objectives.} Recent studies derive token weights from effects on correct-response likelihood~\citep{ren2026effect,deng2026token}, discriminative gradient information~\citep{zhang2026delta}, and the gradient utility of supervised losses~\citep{gong2026vcore}. FACTOR~\citep{ma2026much} uses feedback-conditioned likelihood differences to distribute action-level credit among tokens. Continuous relaxations~\citep{jang2016categorical,maddison2016concrete} and straight-through (ST) estimation~\citep{bengio2013estimating} provide surrogate gradients for discrete decisions. Using surrogate reward gradients, DATC estimates first-order changes in reward along signed GRPO update directions and allocates token-level advantages accordingly.

\section{Method}
\label{sec:method}

DATC (Direction-Aware Token Credit Assignment) evaluates the local reward utility of updates at each position through the inner product between the reward-improvement direction and GRPO's signed update direction. It uses this utility to construct token credit and allocate advantages. The reward-improvement direction is provided by the end-to-end actor--RM surrogate gradient.

\subsection{Displacement and Force in Logit Space}
\label{sec:method:directions}

Our objective is to improve the RM reward of actor-generated responses, namely to maximize $J(\theta)=\mathbb E_{x\sim\mathcal D,\,y\sim\pi_\theta(\cdot\mid x)}[R_\phi(x,y)]$. We construct an end-to-end surrogate loss from the negative RM reward to characterize the reward-improvement direction in the actor's logit space. Let $\widetilde E_\theta$ denote the input embeddings connecting the actor's outputs to the RM. This loss is

\begin{equation}
\mathcal L_{\mathrm{e2e}}(\theta;x,y)
=-R_\phi(x,\widetilde E_\theta).
\tag{1}
\label{eq:1}
\end{equation}

The key to this connection is discrete sampling. The actor outputs a probability distribution over candidate tokens, whereas the RM receives specific tokens sampled from it. We use a straight-through (ST) surrogate~\citep{bengio2013estimating} to specify a usable backward rule for this path while retaining the score of the actual response. Consider a sampled response $y=(a_1,\ldots,a_T)$. Given its sampled prefix $a_{<t}$, the policy distribution at position $t$ is $p_t=\pi_\theta(\cdot\mid x,a_{<t})=\operatorname{softmax}(\ell_t)$. The actor and RM use the same tokenizer with aligned token IDs, establishing a one-to-one correspondence between dimensions of the candidate-token probability vector and rows of the RM embedding matrix. For the sampled token's embedding $e_t$, define\footnote{$W\in\mathbb R^{V\times D}$ is the RM input embedding matrix, $\operatorname{sg}$ denotes stop-gradient, and $\widetilde E_\theta=(\widetilde e_1,\ldots,\widetilde e_T)$.}

\begin{equation}
\widetilde e_t
=W^\top p_t+\operatorname{sg}(e_t-W^\top p_t).
\tag{2}
\label{eq:2}
\end{equation}

In the forward pass, $\widetilde e_t=e_t$, so Equation~\eqref{eq:1} uses the RM score of the actual generated response. In the backward pass, the probability branch maps a change $\delta p_t$ to an input perturbation $W^\top\delta p_t$, propagating the reward's response to its inputs back to the actor's outputs through a differentiable surrogate path. RM parameters remain frozen throughout. Applying the chain rule along this connection and differentiating in logit space yields

\[
-\widehat\nabla_{\ell_t}\mathcal L_{\mathrm{e2e}}
=\widehat\nabla_{\ell_t}R
=\left(\frac{\partial p_t}{\partial\ell_t}\right)^\top
W\nabla_{e_t}R.
\]

The input gradient is evaluated at the actual sampled response, and the hat denotes the surrogate derivative specified by Equation~\eqref{eq:2}. The sampled response and its prefixes remain fixed during differentiation.

\textbf{Displacement in logit space.} With reward improvement as the objective, the displacement points toward higher reward in logit space. Consider a group of responses sampled by GRPO~\citep{shao2024deepseekmath} for the same input. For one response, let $r$ be the response reward used for policy optimization and $\bar r$ the group mean. GRPO computes the advantage $A=(r-\bar r)/\sigma$ using a reward normalization scale $\sigma$; Appendix~\ref{appendix:training:objective} describes reward processing in our experiments. We use the same $\sigma$ to standardize the reward response, writing $g_t=\sigma^{-1}\widehat\nabla_{\ell_t}R$.\footnote{The group mean and reward normalization scale are treated as constants when constructing credit.} This gives the direction $g_t$ for increasing RM reward in logit space. It describes how the logits of candidate tokens in the vocabulary can change to achieve higher local reward under this surrogate. The full theoretical analysis is given in Appendix~\ref{appendix:theory:end-to-end-gradient}.

\textbf{Force in logit space.} The response-level advantage determines whether to reinforce or suppress a sampled token, corresponding in the analogy to the force that drives changes in the model's outputs. Differentiating the sampled token's log probability with respect to the logits gives its score direction, $k_t=\nabla_{\ell_t}\log p_t(a_t)$. Thus, $k_t$ is the direction that increases the sampled token's log probability, while the sign of $A$ determines whether GRPO reinforces along this direction or suppresses in the opposite direction.

\textbf{Work in logit space.} In a local model with independent logits, original GRPO uses the same advantage coefficient for all tokens within a response and averages the policy terms across positions. Its policy gradient step in logit space is $\delta\ell_t=\eta A k_t/T$.\footnote{Here the importance ratio is $1$, and clipping is locally in its linear regime.} Substituting this step into the first-order reward expansion gives

\begin{equation}
\widehat{\Delta(R/\sigma)}
:=\sum_t g_t^\top\delta\ell_t
=\frac\eta T\sum_t A\underbrace{g_t^\top k_t}_{d_t}.
\tag{3}
\label{eq:3}
\end{equation}

This yields directional credit $d_t$. It measures the agreement between the reward direction and the sampled-action direction, accounting for their magnitudes as well. When $d_t>0$, increasing the sampled token's log probability has a positive local reward effect; when $d_t<0$, the opposite holds. The product $A d_t$ further accounts for the actual direction of reinforcement or suppression, indicating how this GRPO update is predicted to affect reward.

\subsection{Aggregating Local Advantages along the Sampled-Action Direction}
\label{sec:method:advantages}

We now interpret directional credit through the reward advantages of candidate actions. The end-to-end surrogate gradient can be decomposed exactly into a weighted sum of candidate actions' score gradients. Define the centered coefficient of the reward input gradient along a candidate token's embedding direction as\footnote{$W_b$ denotes the RM embedding of candidate token $b$, and the input gradient is evaluated at the actual sampled response.}

\begin{equation}
C_t(b)=\left\langle
\nabla_{e_t}R,\,
W_b-\sum_c p_t(c)W_c
\right\rangle.
\tag{4}
\label{eq:4}
\end{equation}

$C_t(b)$ describes the reward response of a candidate token relative to the current probability-weighted average embedding. In the RM's first-order linear reward model, $C_t(b)/\sigma$ is the candidate token's standardized advantage relative to the current policy's average reward, which we call the \textbf{local reward advantage}. Its derivation is given in Appendix~\ref{appendix:theory:end-to-end-gradient}. Using the chain rule, $\nabla p=p\nabla\log p$, and probability normalization, the descent direction of the end-to-end loss can be written as

\begin{equation}
\begin{aligned}
-\frac1\sigma\widehat\nabla_\theta\mathcal L_{\mathrm{e2e}}
=\frac1\sigma\widehat\nabla_\theta R
=\sum_{t,b}p_t(b)
\underbrace{\frac{C_t(b)}{\sigma}}_{\text{local reward advantage}}
\nabla_\theta\log p_t(b).
\end{aligned}
\tag{5}
\label{eq:5}
\end{equation}

Equation~\eqref{eq:5} gives a policy gradient decomposition of the end-to-end surrogate gradient, with the same advantage-weighted structure as REINFORCE~\citep{williams1992simple}. Here the action advantages are defined by the RM's local linear reward model. To see how these advantages relate to $d_t$, we first differentiate with respect to the logits at position $t$ along the same computational path:

\[
\begin{aligned}
g_t
=\frac1\sigma\sum_b C_t(b)\nabla_{\ell_t}p_t(b)
=\sum_b p_t(b)\frac{C_t(b)}{\sigma}
\nabla_{\ell_t}\log p_t(b).
\end{aligned}
\]

Centering preserves the gradient because $\sum_b\nabla_{\ell_t}p_t(b)=0$. Both the parameter-gradient and logit-gradient expressions follow from this chain-rule expansion. Substituting the expression above into $d_t=\langle g_t,k_t\rangle$ and using the linearity of the inner product gives

\begin{equation}
\begin{aligned}
d_t
=\left\langle
\sum_b p_t(b)\frac{C_t(b)}{\sigma}
\nabla_{\ell_t}\log p_t(b),\,k_t
\right\rangle
=\sum_b p_t(b)\frac{C_t(b)}{\sigma}
\left\langle
\nabla_{\ell_t}\log p_t(b),\,
\nabla_{\ell_t}\log p_t(a_t)
\right\rangle.
\end{aligned}
\tag{6}
\label{eq:6}
\end{equation}

Directional credit thus aggregates local reward advantages in the end-to-end gradient: each candidate token's advantage contributes to the final $d_t$ through the relationship between its score direction and the sampled-action direction. DATC thereby converts end-to-end reward information into position-level credit for GRPO. Competition among candidate tokens under the current policy is also encoded in these directions. Expanding the expression yields

\begin{equation}
d_t=\frac1\sigma\left[
p_t(a_t)C_t(a_t)-\sum_b p_t(b)^2C_t(b)
\right].
\tag{7}
\label{eq:7}
\end{equation}

The second term arises from the adjustment of all candidate probabilities induced by softmax normalization. The full derivation and its relationship to shared-parameter updates are given in Appendix~\ref{appendix:theory:grpo-direction}.

\subsection{Converting Directional Gains into Token-Level Advantages}
\label{sec:method:allocation}

Equation~\eqref{eq:3} connects each position's update to the reward objective. We use this connection to allocate token weights $w_t$ while keeping their within-response mean at $1$. Each position thus continues to learn along GRPO's sampled-action direction; the allocation changes its share of the update.

In the same local model, the weighted update is $\delta\ell_t(w)=\eta A w_tk_t/T$, with predicted standardized reward increment $\eta A\sum_t w_td_t/T$. To assign larger shares to positions with higher gains while controlling the allocation's deviation from original GRPO, we solve

\begin{equation}
\max_{\substack{w_t\ge0\\\sum_t w_t=T}}
\left\{
\frac AT\sum_t w_td_t
-\frac\tau T\sum_t w_t\log w_t
\right\},\qquad\tau>0.
\tag{8}
\label{eq:8}
\end{equation}

Here $w_t/T$ is the share assigned to a position, and original GRPO assigns $1/T$ to every position. The regularizer is exactly the KL divergence between these two distributions of shares. Equation~\eqref{eq:8} has a unique optimum, which directly gives the training weights and token-level advantages:

\begin{equation}
\boxed{
w_t^*(\tau)=T\,\operatorname{softmax}_t(A d/\tau),
\qquad
\widetilde A_t=A w_t^*(\tau).
}
\tag{9}
\label{eq:9}
\end{equation}

When $A>0$, the allocation favors positions with larger $d_t$. When $A<0$, it favors positions with smaller $d_t$, directing more suppression toward these tokens. The weights remain positive and have mean $1$, preserving the sign and mean scale of the response-level advantage.

\paragraph{Local gain property.}
For a fixed response and its $A,d$, at any finite $\tau>0$, the predicted gain of Equation~\eqref{eq:9} over original GRPO satisfies

\begin{equation}
\widehat{\Delta(R/\sigma)}(w^*)
-\widehat{\Delta(R/\sigma)}(\mathbf1)
\ge\frac{\eta\tau}{T}\sum_t w_t^*\log w_t^*\ge0.
\tag{10}
\label{eq:10}
\end{equation}

The original weights $\mathbf1$ are feasible in Equation~\eqref{eq:8}, so the additional predicted gain of the optimal allocation is sufficient to offset its KL cost. This property holds in the first-order surrogate model with independent logits in Equation~\eqref{eq:3}; the proof is in Appendix~\ref{appendix:theory:kl-allocation}. It connects the choice of training weights to local improvement in the end-to-end reward.

Utility magnitudes vary across responses and over training, so a fixed temperature can produce substantially different degrees of concentration. We use a hyperparameter $\lambda>1$ to control the maximum weight of an individual token. Starting from temperature $1$, we choose the smallest temperature satisfying the upper bound for each response:

\begin{equation}
\tau_i=\min\left\{\tau\ge1:
\max_t w_{it}^*(\tau)\le\lambda\right\}.
\tag{11}
\label{eq:11}
\end{equation}

Once $\tau_i$ is selected, the weights remain optimal at the corresponding KL regularization strength, and Equation~\eqref{eq:10} continues to hold. Temperature selection and its feasibility are described in Appendix~\ref{appendix:theory:adaptive-temperature}. The experimental allocator uses within-response standardization and constrains weights to $[1/\lambda,\lambda]$. These operations preserve the same KL-optimal allocation structure, as shown in Appendix~\ref{appendix:theory:practical-allocation}.

Finally, we cache $\widetilde A_t$ as a fixed coefficient and substitute it into GRPO's clipped policy objective, retaining the baseline's reference-policy KL term. One RM input backward pass provides reward gradients at all positions in a response. These gradients, together with old-policy probabilities, determine the weights subsequently used for the actor update. The complete training objective and algorithm are given in Appendix~\ref{appendix:training:overview}.

\section{Experiments}
\label{sec:experiments}

We evaluate DATC on general preference alignment to examine whether token credit assignment based on local work improves policy optimization and generation quality. We consider Base and Instruct initializations in each of two model families, yielding four experimental settings. Method comparisons and component ablations analyze the roles of directional information, the correspondence between weights and positions, and allocation concentration.

\subsection{Experimental Setup}
\label{sec:experiments:setup}

\paragraph{Training setup.}We perform online preference optimization using general-purpose instructions from UltraFeedback~\citep{cui2023ultrafeedback}, with responses sampled by the policy model and scored by a frozen RM. The policy models are Qwen3-14B~\citep{yang2025qwen3} and Llama-3.1-8B~\citep{grattafiori2024llama}, paired with tokenizer-aligned Skywork-Reward-V2-Qwen3-8B~\citep{liu2026skywork} and Skywork-Reward-Llama-3.1-8B-v0.2~\citep{liu2024skywork}, respectively. Each family is evaluated with two initializations: reinforcement learning starts from a model after lightweight SFT in the Base setting and from the official instruction-tuned model in the Instruct setting. Within each setting, DATC and GRPO share the training initialization, rollout budget, reward processing, and LoRA~\citep{hu2021lora}  optimization settings, differing only in token-level advantage allocation.

\paragraph{Evaluation protocol.}We evaluate the trained policies using four metrics. RM-Reward is the mean raw score assigned by the frozen training RM on independent held-out prompts, with comparisons restricted to policies evaluated by the same RM.
AlpacaEval~\citep{li2023alpacaeval} and Arena-Hard v2.0~\citep{li2024crowdsourced} report preference win rates against reference responses on general open-ended instructions and challenging real-user prompts, respectively. MT-Bench~\citep{zheng2023judging} reports the mean score of multi-turn responses across tasks such as writing, reasoning, and coding to assess overall response quality. Win rates are reported as percentages and their differences as percentage points. Training and evaluation details and supplementary analyses are provided in Appendix~\ref{appendix:experiments:overview}.

\subsection{Overall Performance and Training Dynamics}
\label{sec:experiments:overall}

Table~\ref{tab:main-results} compares DATC and GRPO across the two model families and two initialization settings. DATC uses $\lambda=4$. Initial and SFT denote the original checkpoint and the Base checkpoint after lightweight SFT, respectively. AlpacaEval reports raw win rates. Bold indicates the best result within each experimental setting.

\begin{table}[t]
\caption{Main experimental results, with DATC reported as mean ± standard deviation across three random seeds.}
\label{tab:main-results}
\centering
\newcommand{\datcstd}[2]{\textbf{#1}\,{\color{black!55}$\pm$\,#2}}
\begin{tabularx}{\linewidth}{@{}Xrrrr@{}}
\toprule
Method & RM-Reward $\uparrow$ & AlpacaEval $\uparrow$ & Arena-Hard $\uparrow$ & MT-Bench $\uparrow$ \\
\midrule
\multicolumn{5}{@{}l}{\textbf{Qwen3-14B-Base}} \\
Initial & -2.54 & 6.65 & 8.42 & 6.76 \\
SFT & 3.75 & 7.00 & 3.15 & 6.53 \\
GRPO & 10.87 & 37.39 & 28.49 & 7.70 \\
DATC & \datcstd{13.35}{0.16} & \datcstd{49.57}{1.12} & \datcstd{39.31}{1.79} & \datcstd{7.99}{0.05} \\
\midrule
\multicolumn{5}{@{}l}{\textbf{Qwen3-14B-Instruct}} \\
Initial & 11.43 & 63.35 & 66.75 & 7.96 \\
GRPO & 13.75 & 72.80 & 73.60 & 8.36 \\
DATC & \datcstd{14.43}{0.09} & \datcstd{73.66}{1.59} & \datcstd{74.86}{1.43} & \datcstd{8.41}{0.09} \\
\midrule
\multicolumn{5}{@{}l}{\textbf{Llama-3.1-8B-Base}} \\
Initial & -28.06 & 0.00 & 0.03 & 2.01 \\
SFT & -8.08 & 4.02 & 1.24 & 4.91 \\
GRPO & 4.55 & 9.73 & 2.83 & 5.61 \\
DATC & \datcstd{11.80}{0.39} & \datcstd{12.58}{0.85} & \datcstd{5.10}{0.48} & \datcstd{5.78}{0.06} \\
\midrule
\multicolumn{5}{@{}l}{\textbf{Llama-3.1-8B-Instruct}} \\
Initial & -0.26 & 11.47 & 3.02 & 6.81 \\
GRPO & 19.61 & 21.61 & 8.41 & 7.13 \\
DATC & \datcstd{24.73}{1.04} & \datcstd{27.08}{1.25} & \datcstd{9.35}{0.63} & \datcstd{7.45}{0.05} \\
\bottomrule
\end{tabularx}
\end{table}

Across the Base and Instruct settings of Qwen3-14B and Llama-3.1-8B, DATC achieves higher mean scores than the corresponding GRPO baseline on all four evaluation metrics. For Qwen3-14B-Base, the AlpacaEval and Arena-Hard win rates improve by 12.18 and 10.82 percentage points, respectively.

\begin{figure}[t]
\centering
\includegraphics[width=\linewidth]{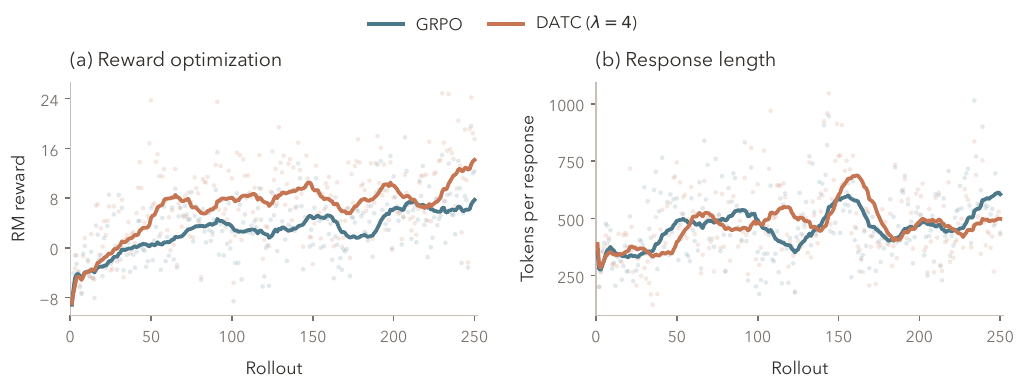}
\caption{Raw RM reward (left) and mean response length (right) for Llama-3.1-8B-Base after lightweight SFT. Light-colored points indicate the mean for each rollout; solid lines show the trailing 25-rollout moving average.}
\label{fig:training-main}
\end{figure}

Figure~\ref{fig:training-main} shows training dynamics for Llama-3.1-8B-Base after lightweight SFT. DATC's raw RM reward trend is above GRPO's during most of training, with mean rewards of 14.14 and 7.72, respectively, over the final 25 rollouts. DATC also produces shorter responses in this final window, averaging 497.9 tokens versus 605.2 for GRPO. Across the four experimental settings above, DATC increases total GPU time by only 5.7\%--9.1\% relative to GRPO under the same setting and rollout budget.

\subsection{Credit Assignment Comparisons and Ablations}
\label{sec:experiments:ablation}

Table~\ref{tab:credit-ablation} compares GRPO, P2T, RED, DATC, and its ablation variants from the same Qwen3-14B-Base initialization after lightweight SFT, with identical policy optimization settings. The ablations use DATC with $\lambda=4$ as the reference, varying directional information, the correspondence between weights and positions, and the weight upper bound.

\begin{table}[t]
\caption{Method comparisons and credit assignment ablations on Qwen3-14B-Base.}
\label{tab:credit-ablation}
\centering
\begin{tabularx}{\linewidth}{@{}Xrrrr@{}}
\toprule
Method & RM-Reward $\uparrow$ & AlpacaEval $\uparrow$ & Arena-Hard $\uparrow$ & MT-Bench $\uparrow$ \\
\midrule
Initial & -2.54 & 6.65 & 8.42 & 6.76 \\
SFT & 3.75 & 7.00 & 3.15 & 6.53 \\
GRPO & 10.87 & 37.39 & 28.49 & 7.70 \\
P2T & 10.96 & 36.40 & 29.68 & 7.66 \\
RED & 6.14 & 13.21 & 22.30 & 6.93 \\
\midrule
\multicolumn{5}{@{}l}{\textbf{DATC}} \\
\quad $\lambda=4$ & \textbf{13.35} & \textbf{49.57} & \textbf{39.31} & \textbf{7.99} \\
\quad $\lambda=2$ & 12.89 & 41.49 & 34.49 & 7.77 \\
\quad $\lambda=8$ & 12.91 & 45.96 & 37.63 & 7.98 \\
\quad w/o direction alignment & -4.02 & 1.53 & 1.23 & 2.71 \\
\quad w/ shuffled weights & 10.73 & 32.62 & 26.99 & 7.51 \\
\bottomrule
\end{tabularx}
\end{table}

\textbf{Comparison with token reward attribution.} 
RED~\citep{li2025red} redistributes response-level rewards using differences between successive RM prefix scores, while P2T~\citep{wu2026unlocking} uses RM input gradients to estimate token replacement effects. DATC incorporates the current policy’s candidate-token distribution and sampled-action update direction to estimate local reward utility. In this setting, DATC outperforms both RED and P2T on all four evaluation metrics.

\textbf{Role of direction alignment.} The direction-alignment ablation (w/o direction alignment) removes the angle term from the credit inner product, retaining only the product of the two direction norms. Relative to full DATC ($\lambda=4$), its RM-Reward decreases to $-4.02$, while AlpacaEval and Arena-Hard win rates decrease to 1.53\% and 1.23\%, respectively. This indicates that direction alignment is a key factor in DATC's performance gains in this setting.

\textbf{Shuffled-weight variant.} The shuffled-weight variant (w/ shuffled weights) permutes DATC weights within each response, preserving the weight set and its concentration while changing only the correspondence between weights and token positions. Relative to full DATC, AlpacaEval and Arena-Hard decrease by 16.95 and 12.32 percentage points, respectively, and MT-Bench decreases by 0.48 points. Performance gains therefore also depend on matching weights to specific update positions.

\textbf{Allocation concentration.} The weight upper bound $\lambda$ controls the maximum update share an individual token can receive. Among $\lambda\in\{2,4,8\}$, $\lambda=4$ achieves the highest mean on all four metrics.

\subsection{Token-Level Weight Analysis}
\label{sec:experiments:cases}

\begin{figure}[t]
\centering
\includegraphics[width=\linewidth]{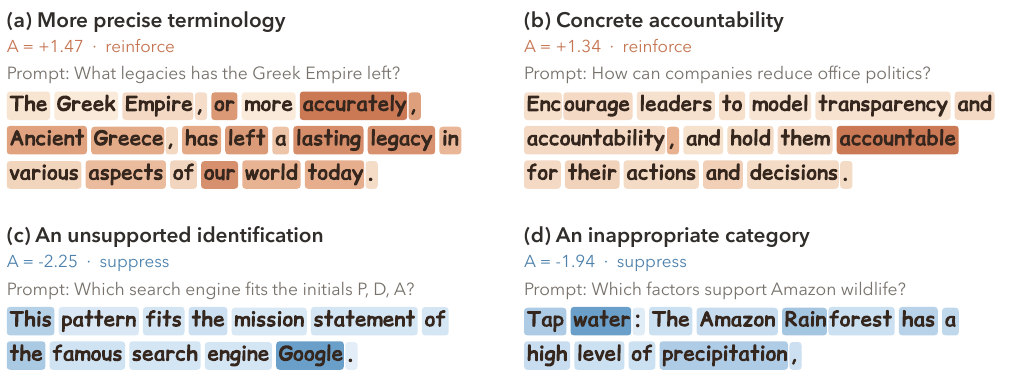}
\caption{DATC weights for excerpts from actual training responses. Warm orange denotes reinforcement and blue denotes suppression; darker shades within each color indicate larger weights.}
\label{fig:token-cases}
\end{figure}

Figure~\ref{fig:token-cases} shows the actual training weights for four response excerpts. With a positive advantage, a higher weight increases reinforcement; with a negative advantage, it increases suppression. In the positive-advantage cases, ``accurately'' receives a very high weight when correcting ``Greek Empire'' to ``Ancient Greece.'' Likewise, ``accountable'' receives a very high weight in the actionable advice ``hold them accountable.'' In a response that identifies a search engine from insufficient clues, ``Google'' receives a higher weight, increasing suppression of that inference. In a response that incorrectly lists tropical rainforest precipitation as ``Tap water,'' ``water'' also receives a higher weight. Together, these cases illustrate how DATC assigns different update strengths to specific expressions within the response-level direction of reinforcement or suppression.

\begin{samepage}
\section{Conclusion}
\label{sec:conclusion}

We introduce DATC, which adopts a work-based view of credit to allocate token-level advantages according to the alignment between GRPO updates and the RM reward-improvement direction. Its KL-regularized allocation provides a local reward gain guarantee in the surrogate model. DATC outperforms GRPO across all four settings of two model families.
\par
\end{samepage}

\bibliography{references}

@article{chan2024dense,
  title={Dense reward for free in reinforcement learning from human feedback},
  author={Chan, Alex J and Sun, Hao and Holt, Samuel and Van Der Schaar, Mihaela},
  journal={arXiv preprint arXiv:2402.00782},
  year={2024}
}

@misc{li2023alpacaeval,
  title={Alpacaeval: An automatic evaluator of instruction-following models},
  author={Li, Xuechen and Zhang, Tianyi and Dubois, Yann and Taori, Rohan and Gulrajani, Ishaan and Guestrin, Carlos and Liang, Percy and Hashimoto, Tatsunori B},
  year={2023}
}

@article{li2024crowdsourced,
  title={From crowdsourced data to high-quality benchmarks: Arena-hard and benchbuilder pipeline},
  author={Li, Tianle and Chiang, Wei-Lin and Frick, Evan and Dunlap, Lisa and Wu, Tianhao and Zhu, Banghua and Gonzalez, Joseph E and Stoica, Ion},
  journal={arXiv preprint arXiv:2406.11939},
  year={2024}
}

@article{maddison2016concrete,
  title={The concrete distribution: A continuous relaxation of discrete random variables},
  author={Maddison, Chris J and Mnih, Andriy and Teh, Yee Whye},
  journal={arXiv preprint arXiv:1611.00712},
  year={2016}
}

@article{zhang2026delta,
  title={DelTA: Discriminative Token Credit Assignment for Reinforcement Learning from Verifiable Rewards},
  author={Zhang, Kaiyi and Wu, Wei and Lin, Yankai},
  journal={arXiv preprint arXiv:2605.21467},
  year={2026}
}

@article{ma2026much,
  title={How Much, Then Where: Credit-Conserving Action-to-Token Allocation for Multi-Turn Agent Reinforcement Learning},
  author={Ma, Lichao and Sun, Yang and Zhao, Shuaitao and Fang, Yangyi and Qin, Cong and Fu, Xiaoliang and Tian, Yuhang and Wei, Yuchen and Zhu, Junbo and Wei, Yang and others},
  journal={arXiv preprint arXiv:2608.07118},
  year={2026}
}

@article{pala2026grail,
  title={GRAIL: Gradient-Reweighted Advantages for Reinforcement Learning with Verifiable Rewards},
  author={Pala, Tej Deep and Toh, Vernon and Poria, Soujanya},
  journal={arXiv preprint arXiv:2606.04889},
  year={2026}
}

@article{shao2024deepseekmath,
  title={Deepseekmath: Pushing the limits of mathematical reasoning in open language models},
  author={Shao, Zhihong and Wang, Peiyi and Zhu, Qihao and Xu, Runxin and Song, Junxiao and Bi, Xiao and Zhang, Haowei and Zhang, Mingchuan and Li, YK and Wu, Yang and others},
  journal={arXiv preprint arXiv:2402.03300},
  year={2024}
}

@article{tan2025gtpo,
  title={Gtpo and grpo-s: Token and sequence-level reward shaping with policy entropy},
  author={Tan, Hongze and Wang, Zihan and Pan, Jianfei and Lin, Jinghao and Wang, Hao and Wu, Yifan and Chen, Tao and Zheng, Zhihang and Tang, Zhihao and Yang, Haihua},
  journal={arXiv preprint arXiv:2508.04349},
  year={2025}
}

@article{zhang2026grad2reward,
  title={Grad2reward: From sparse judgment to dense rewards for improving open-ended llm reasoning},
  author={Zhang, Zheng and Lu, Ao and Zeng, Yuanhao and Shan, Ziwei and Guo, Jinjin and Li, Lufei and Li, Yexin and Ren, Kan},
  journal={arXiv preprint arXiv:2602.01791},
  year={2026}
}

@article{jang2016categorical,
  title={Categorical reparameterization with gumbel-softmax},
  author={Jang, Eric and Gu, Shixiang and Poole, Ben},
  journal={arXiv preprint arXiv:1611.01144},
  year={2016}
}

@article{yuan2024free,
  title={Free process rewards without process labels},
  author={Yuan, Lifan and Li, Wendi and Chen, Huayu and Cui, Ganqu and Ding, Ning and Zhang, Kaiyan and Zhou, Bowen and Liu, Zhiyuan and Peng, Hao},
  journal={arXiv preprint arXiv:2412.01981},
  year={2024}
}

@article{grattafiori2024llama,
  title={The llama 3 herd of models},
  author={Grattafiori, Aaron and Dubey, Abhimanyu and Jauhri, Abhinav and Pandey, Abhinav and Kadian, Abhishek and Al-Dahle, Ahmad and Letman, Aiesha and Mathur, Akhil and Schelten, Alan and Vaughan, Alex and others},
  journal={arXiv preprint arXiv:2407.21783},
  year={2024}
}

@article{hu2021lora,
  title={Lora: Low-rank adaptation of large language models},
  author={Hu, Edward J and Shen, Yelong and Wallis, Phillip and Allen-Zhu, Zeyuan and Li, Yuanzhi and Wang, Shean and Wang, Lu and Chen, Weizhu},
  journal={arXiv preprint arXiv:2106.09685},
  year={2021}
}

@article{zheng2023judging,
  title={Judging llm-as-a-judge with mt-bench and chatbot arena},
  author={Zheng, Lianmin and Chiang, Wei-Lin and Sheng, Ying and Zhuang, Siyuan and Wu, Zhanghao and Zhuang, Yonghao and Lin, Zi and Li, Zhuohan and Li, Dacheng and Xing, Eric and others},
  journal={Advances in neural information processing systems},
  volume={36},
  pages={46595--46623},
  year={2023}
}

@article{ren2026effect,
  title={On the effect of negative gradient in group relative deep reinforcement optimization},
  author={Ren, Yi and Li, Muchen and Sutherland, Danica J and Li, Xiaoxiao and Thrampoulidis, Christos and others},
  journal={Advances in Neural Information Processing Systems},
  volume={38},
  pages={73828--73856},
  year={2026}
}

@inproceedings{li2026outcome,
  title={Outcome-Grounded Advantage Reshaping for Fine-Grained Credit Assignment in Mathematical Reasoning},
  author={Li, Ziheng and Kang, Liu and Xiao, Feng and Xing, Luxi and Si, Qingyi and Li, Zhuoran and Gong, Weikang and Yang, Deqing and Xiao, Yanghua and Guo, Hongcheng},
  booktitle={Proceedings of the 64th Annual Meeting of the Association for Computational Linguistics (Volume 1: Long Papers)},
  pages={24681--24693},
  year={2026}
}

@inproceedings{wu2026unlocking,
  title={Unlocking Token Rewards via Training-Free Reward Attribution},
  author={Wu, Sitong and Tan, Haoru and Xia, Bin and Zhang, Xichen and Li, Jingyao and Zhang, Shaofeng and Qi, Xiaojuan and Yu, Bei and Jia, Jiaya},
  booktitle={Proceedings of the IEEE/CVF Conference on Computer Vision and Pattern Recognition},
  pages={5082--5091},
  year={2026}
}

@article{chen2025discriminative,
  title={Discriminative policy optimization for token-level reward models},
  author={Chen, Hongzhan and Yang, Tao and Gao, Shiping and Chen, Ruijun and Quan, Xiaojun and Tian, Hongtao and Yao, Ting},
  journal={arXiv preprint arXiv:2505.23363},
  year={2025}
}

@article{yang2025qwen3,
  title={Qwen3 technical report},
  author={Yang, An and Li, Anfeng and Yang, Baosong and Zhang, Beichen and Hui, Binyuan and Zheng, Bo and Yu, Bowen and Gao, Chang and Huang, Chengen and Lv, Chenxu and others},
  journal={arXiv preprint arXiv:2505.09388},
  year={2025}
}

@inproceedings{li2025red,
  title={RED: Unleashing token-level rewards from holistic feedback via reward redistribution},
  author={Li, Jiahui and Li, Lin and Chang, Tai-Wei and Kuang, Kun and Chen, Long and Zhou, Jun and Yang, Cheng},
  booktitle={Proceedings of the 2025 Conference on Empirical Methods in Natural Language Processing},
  pages={4993--5022},
  year={2025}
}

@article{williams1992simple,
  title={Simple statistical gradient-following algorithms for connectionist reinforcement learning},
  author={Williams, Ronald J},
  journal={Machine learning},
  volume={8},
  number={3},
  pages={229--256},
  year={1992},
  publisher={Springer}
}

@inproceedings{setlur2025rewarding,
  title={Rewarding progress: Scaling automated process verifiers for llm reasoning},
  author={Setlur, Amrith and Nagpal, Chirag and Fisch, Adam and Geng, Xinyang and Eisenstein, Jacob and Agarwal, Rishabh and Agarwal, Alekh and Berant, Jonathan and Kumar, Aviral},
  booktitle={International Conference on Learning Representations},
  volume={2025},
  pages={60808--60838},
  year={2025}
}

@article{guo2026segment,
  title={Segment policy optimization: Effective segment-level credit assignment in rl for large language models},
  author={Guo, Yiran and Xu, Lijie and Liu, Jie and Ye, Dan and Qiu, Shuang},
  journal={Advances in Neural Information Processing Systems},
  volume={38},
  pages={114399--114431},
  year={2026}
}

@article{bengio2013estimating,
  title={Estimating or propagating gradients through stochastic neurons for conditional computation},
  author={Bengio, Yoshua and L{\'e}onard, Nicholas and Courville, Aaron},
  journal={arXiv preprint arXiv:1308.3432},
  year={2013}
}

@article{liu2024skywork,
  title={Skywork-reward: Bag of tricks for reward modeling in llms},
  author={Liu, Chris Yuhao and Zeng, Liang and Liu, Jiacai and Yan, Rui and He, Jujie and Wang, Chaojie and Yan, Shuicheng and Liu, Yang and Zhou, Yahui},
  journal={arXiv preprint arXiv:2410.18451},
  year={2024}
}

@inproceedings{liu2026skywork,
  title={Skywork-reward-v2: Scaling preference data curation via human-ai synergy},
  author={Liu, Yuhao and Zeng, Liang and Xiao, Yuzhen and He, Jujie and Liu, Jiacai and Wang, Chaojie and Yan, Rui and Shen, Wei and Zhang, Fuxiang and Xu, Jiacheng and others},
  booktitle={International Conference on Learning Representations},
  volume={2026},
  pages={133805--133838},
  year={2026}
}

@inproceedings{lightman2024let,
  title={Let's verify step by step},
  author={Lightman, Hunter and Kosaraju, Vineet and Burda, Yuri and Edwards, Harrison and Baker, Bowen and Lee, Teddy and Leike, Jan and Schulman, John and Sutskever, Ilya and Cobbe, Karl},
  booktitle={International Conference on Learning Representations},
  volume={2024},
  pages={39578--39601},
  year={2024}
}

@inproceedings{deng2026token,
  title={Token hidden reward: Steering exploration-exploitation in group relative deep reinforcement learning},
  author={Deng, Wenlong and Ren, Yi and Li, Yushu and Gong, Boying and Sutherland, Danica and Li, Xiaoxiao and Thrampoulidis, Christos},
  booktitle={International Conference on Learning Representations},
  volume={2026},
  pages={122109--122133},
  year={2026}
}

@article{cui2023ultrafeedback,
  title={Ultrafeedback: Boosting language models with scaled ai feedback},
  author={Cui, Ganqu and Yuan, Lifan and Ding, Ning and Yao, Guanming and He, Bingxiang and Zhu, Wei and Ni, Yuan and Xie, Guotong and Xie, Ruobing and Lin, Yankai and others},
  journal={arXiv preprint arXiv:2310.01377},
  year={2023}
}

@inproceedings{gong2026vcore,
  title={VCORE: Variance-Controlled Optimization-based Reweighting for Chain-of-Thought Supervision},
  author={Gong, Xuan and Wang, Senmiao and Huang, Hanbo and Sun, Ruoyu and Liang, Shiyu},
  booktitle={Proceedings of the 64th Annual Meeting of the Association for Computational Linguistics (Volume 1: Long Papers)},
  pages={28161--28180},
  year={2026}
}

@article{kazemnejad2024vineppo,
  title={Vineppo: Refining credit assignment in rl training of llms},
  author={Kazemnejad, Amirhossein and Aghajohari, Milad and Portelance, Eva and Sordoni, Alessandro and Reddy, Siva and Courville, Aaron and Roux, Nicolas Le},
  journal={arXiv preprint arXiv:2410.01679},
  year={2024}
}

@inproceedings{wang2026emergent,
  title={Emergent hierarchical reasoning in llms through reinforcement learning},
  author={Wang, Haozhe and Xu, Qixin and Liu, Che and Wu, Junhong and Lin, Fangzhen and Chen, Wenhu},
  booktitle={International Conference on Learning Representations},
  volume={2026},
  pages={74519--74543},
  year={2026}
}

@article{zhong2024dpo,
  title={Dpo meets ppo: Reinforced token optimization for rlhf},
  author={Zhong, Han and Shan, Zikang and Feng, Guhao and Xiong, Wei and Cheng, Xinle and Zhao, Li and He, Di and Bian, Jiang and Wang, Liwei},
  journal={arXiv preprint arXiv:2404.18922},
  year={2024}
}

@article{cui2025process,
  title={Process reinforcement through implicit rewards},
  author={Cui, Ganqu and Yuan, Lifan and Wang, Zefan and Wang, Hanbin and Zhang, Yuchen and Chen, Jiacheng and Li, Wendi and He, Bingxiang and Fan, Yuchen and Yu, Tianyu and others},
  journal={arXiv preprint arXiv:2502.01456},
  year={2025}
}

@article{hurst2024gpt,
  title={Gpt-4o system card},
  author={Hurst, Aaron and Lerer, Adam and Goucher, Adam P and Perelman, Adam and Ramesh, Aditya and Clark, Aidan and Ostrow, AJ and Welihinda, Akila and Hayes, Alan and Radford, Alec and others},
  journal={arXiv preprint arXiv:2410.21276},
  year={2024}
}

@article{achiam2023gpt,
  title={Gpt-4 technical report},
  author={Achiam, Josh and Adler, Steven and Agarwal, Sandhini and Ahmad, Lama and Akkaya, Ilge and Aleman, Florencia Leoni and Almeida, Diogo and Altenschmidt, Janko and Altman, Sam and Anadkat, Shyamal and others},
  journal={arXiv preprint arXiv:2303.08774},
  year={2023}
}
\bibliographystyle{plainnat}

\clearpage
\appendix
\numberwithin{equation}{section}
\numberwithin{figure}{section}
\numberwithin{table}{section}
\renewcommand{\theequation}{\thesection\arabic{equation}}
\renewcommand{\thefigure}{\thesection\arabic{figure}}
\renewcommand{\thetable}{\thesection\arabic{table}}
\renewcommand{\theHequation}{\thesection.\arabic{equation}}
\renewcommand{\theHfigure}{\thesection.\arabic{figure}}
\renewcommand{\theHtable}{\thesection.\arabic{table}}

\section{Derivations and Theoretical Properties}
\label{appendix:theory:overview}

Following the derivation in the main text, this appendix explains how the end-to-end surrogate gradient yields directional credit and why credit allocation can improve the local predicted gain. We then extend these results to the standardization and weight range used in the experiments.

\subsection{End-to-End Surrogate Gradients at the Hard Response}
\label{appendix:theory:end-to-end-gradient}

The main text starts from the ST end-to-end loss, obtains a logit update direction, and then evaluates the GRPO update. This section expands the surrogate derivative of that loss, proves the centering and policy gradient decomposition in Equations~\eqref{eq:4} and~\eqref{eq:5}, and explains the meaning of local reward advantages.

For the hard response $y$ from a rollout, fix the prefix $s_t=(x,a_{<t})$ at each position. Let $p_t(\theta)=\pi_\theta(\cdot\mid s_t)$, with all derivatives evaluated at $\theta_{\mathrm{old}}$. The group mean and reward normalization scale are treated as constants. Let $r$ be the response reward used for policy optimization. For responses scored according to Equation~\eqref{eq:B1}, the length-dependent reward terms are also constant once the response is fixed, so $(r-\bar r)/\sigma$ and $R/\sigma$ have the same surrogate derivative with respect to the actor.

RM parameters $\phi$, including the input embedding matrix $W$, remain frozen during differentiation. Let $E=(e_1,\ldots,e_T)$ be the actual input embeddings of the response tokens. The prompt input is represented by $x$ and remains fixed.

For a fixed sampled response, the loss in Equation~\eqref{eq:1} is

\[
\mathcal L_{\mathrm{e2e}}(\theta;x,y)
=-R_\phi(x,\widetilde E_\theta).
\]

Equation~\eqref{eq:2} returns $e_t$ in the forward pass and differentiates through the probability branch in the backward pass. The surrogate direction for minimizing this loss is therefore $-\widehat\nabla_\theta\mathcal L_{\mathrm{e2e}}=\widehat\nabla_\theta R$. Applying the product rule along the probability branch gives

\begin{equation}
\widehat d\widetilde e_t
=d(W^\top p_t)
=W^\top dp_t+(dW^\top)p_t
=W^\top dp_t.
\tag{A1}
\label{eq:A1}
\end{equation}

The last equality uses the frozen-parameter condition $dW=0$. The input derivative retained after freezing propagates changes in actor probabilities to the RM scoring function.

To simplify the following calculations, define

\begin{equation}
f_t=\left.\nabla_{\widetilde e_t}R_\phi(x,\widetilde E)\right|_{\widetilde E=E},
\qquad
v_t=Wf_t.
\tag{A2}
\label{eq:A2}
\end{equation}

Here $f_t$ is the gradient at the actual input, and $v_t(b)=\langle f_t,W_b\rangle$ is its coefficient along the embedding direction of candidate token $b$. The chain rule gives

\begin{equation}
\begin{aligned}
\widehat\nabla_\theta R
&=\sum_t\left(\frac{\partial p_t}{\partial\theta}\right)^\top Wf_t\\
&=\sum_{t,b}v_t(b)\nabla_\theta p_t(b)\\
&=\sum_{t,b}p_t(b)v_t(b)\nabla_\theta\log p_t(b).
\end{aligned}
\tag{A3}
\label{eq:A3}
\end{equation}

Let $\mu_t=\sum_b p_t(b)v_t(b)$. Probability normalization implies

\[
\sum_b p_t(b)\mu_t\nabla_\theta\log p_t(b)
=\mu_t\nabla_\theta\sum_b p_t(b)=0.
\]

We can therefore subtract the same $\mu_t$ from every candidate-token coefficient. Define

\begin{equation}
C_t(b)=v_t(b)-\mu_t
=\left\langle f_t,W_b-\sum_c p_t(c)W_c\right\rangle,
\qquad
\sum_b p_t(b)C_t(b)=0.
\tag{A4}
\label{eq:A4}
\end{equation}

Substituting into Equation~\eqref{eq:A3} yields the policy gradient decomposition of the end-to-end surrogate gradient:

\begin{equation}
\frac1\sigma\widehat\nabla_\theta R
=\sum_{t,b}p_t(b)\frac{C_t(b)}{\sigma}
\nabla_\theta\log p_t(b).
\tag{A5}
\label{eq:A5}
\end{equation}

$p_t(b)$ is the candidate action's probability weight, and $C_t(b)/\sigma$ is the coefficient of its score gradient. Equation~\eqref{eq:A5} is an algebraic identity at the fixed expansion point, with the same weighting structure as REINFORCE. All candidate coefficients are obtained from the input gradient at the same hard response.

Take a first-order expansion of the RM at the actual input $E$. Holding all other positions fixed and substituting the embedding of candidate token $b$ into this linear model gives

\begin{equation}
R_t^{\mathrm{lin}}(b)
=R_\phi(x,E)+f_t^\top(W_b-e_t).
\tag{A5a}
\label{eq:A5a}
\end{equation}

Here $E$, $e_t$, and $f_t$ all take their fixed values at the actual sampled response. Under this linear model, the candidate token's standardized advantage relative to the current policy's average reward is

\begin{equation}
\begin{aligned}
\frac{R_t^{\mathrm{lin}}(b)-\sum_c p_t(c)R_t^{\mathrm{lin}}(c)}{\sigma}
&=\frac1\sigma f_t^\top\left(W_b-\sum_c p_t(c)W_c\right)\\
&=\frac{C_t(b)}{\sigma}.
\end{aligned}
\tag{A5b}
\label{eq:A5b}
\end{equation}

Thus, $C_t(b)/\sigma$ is the standardized action advantage in the RM's local linear reward model, referred to as the local reward advantage in the main text. Its probability-weighted mean under the current policy is zero. The exactness of Equation~\eqref{eq:A5} concerns the identity decomposing the ST surrogate gradient, while the reward interpretation of these advantages is determined by the local linear model in Equation~\eqref{eq:A5a}.

Differentiating with respect to the logits at position $t$ along the same end-to-end path gives

\[
-\widehat\nabla_{\ell_t}\mathcal L_{\mathrm{e2e}}
=\widehat\nabla_{\ell_t}R
=\left(\frac{\partial p_t}{\partial\ell_t}\right)^\top Wf_t.
\]

The softmax Jacobian is $\operatorname{diag}(p_t)-p_tp_t^\top$. Using the same normalization scale as the GRPO advantage yields

\begin{equation}
g_t
=\frac1\sigma\widehat\nabla_{\ell_t}R
=\frac1\sigma
\left[\operatorname{diag}(p_t)-p_tp_t^\top\right]v_t
=\frac1\sigma p_t\odot C_t.
\tag{A6}
\label{eq:A6}
\end{equation}

With $p_t^{\mathrm{old}}$ and $e_t$ fixed at the expansion point, Equation~\eqref{eq:A1} can also be obtained from the following continuous local extension:

\begin{equation}
e_t^{\mathrm{loc}}(\ell_t)
=e_t+W^\top\left[p_t(\ell_t)-p_t^{\mathrm{old}}\right].
\tag{A7}
\label{eq:A7}
\end{equation}

It equals the hard input $e_t$ at the old policy, and its ordinary derivative is exactly the Jacobian specified by ST. Passing these inputs to the RM defines a standardized scoring function with gradient $g_t$ at the expansion point. If this local function is twice continuously differentiable, its finite perturbation can be written as

\begin{equation}
\frac{
R_\phi(x,E^{\mathrm{loc}}(\ell+\delta\ell))
-R_\phi(x,E)}
{\sigma}
=
\sum_t g_t^\top\delta\ell_t
+O(\|\delta\ell\|^2).
\tag{A8}
\label{eq:A8}
\end{equation}

The hatted increment in the main text denotes the first-order term. Equation~\eqref{eq:A5} is an identity decomposing the specified surrogate gradient, while the expected reward $J(\theta)$ under discrete generation is the overall training objective. They are connected through the policy optimization procedure in Appendix~\ref{appendix:training:overview}.

\subsection{From the GRPO Direction to Token Credit}
\label{appendix:theory:grpo-direction}

The end-to-end gradient describes the reward response across candidate tokens, whereas GRPO updates the policy through the sampled token's log probability. Substituting the latter into the former's first-order expansion yields the directional credit in the main text. We further show below how local reward advantages enter position-level allocation through the inner product of these two directions.

At position $t$,

\[
\log p_t(a_t)=\ell_{t,a_t}-\log\sum_b\exp(\ell_{t,b}),
\]

and hence

\begin{equation}
k_t(b)
=\frac{\partial\log p_t(a_t)}{\partial\ell_{t,b}}
=\mathbf1\{b=a_t\}-p_t(b).
\tag{A9}
\label{eq:A9}
\end{equation}

Let $\rho_t(\theta)=\pi_\theta(a_t\mid s_t)/\pi_{\mathrm{old}}(a_t\mid s_t)$. At the old policy, $\rho_t=1$, and the policy term in a neighborhood where clipping is inactive is $A\rho_t/T$. Its logit derivative is $A k_t/T$. With cached weights $w_t$, the gradient ascent step in independent logit coordinates becomes

\begin{equation}
\delta\ell_t(w)=\frac{\eta A}{T}w_tk_t.
\tag{A10}
\label{eq:A10}
\end{equation}

Substituting into the first-order term of Equation~\eqref{eq:A8} gives

\begin{equation}
\widehat{\Delta(R/\sigma)}(w)
=\frac{\eta A}{T}\sum_t w_t
\underbrace{g_t^\top k_t}_{d_t}.
\tag{A11}
\label{eq:A11}
\end{equation}

Setting $w_t=1$ recovers Equation~\eqref{eq:3} in the main text. The marginal predicted gain per unit weight is $\eta A d_t/T$; $d_t$ itself describes the rate of reward change along the sampled action's score direction.

To substitute the action advantages from the policy gradient decomposition into directional credit, first compute the logit gradient at position $t$ along the same computational path:

\begin{equation}
\begin{aligned}
g_t
&=\frac1\sigma\sum_b v_t(b)\nabla_{\ell_t}p_t(b)\\
&=\frac1\sigma\sum_b C_t(b)\nabla_{\ell_t}p_t(b)\\
&=\sum_b p_t(b)\frac{C_t(b)}{\sigma}
\nabla_{\ell_t}\log p_t(b).
\end{aligned}
\tag{A12}
\label{eq:A12}
\end{equation}

The second equality uses $C_t(b)=v_t(b)-\mu_t$ and $\sum_b\nabla_{\ell_t}p_t(b)=0$, and the third uses $\nabla p=p\nabla\log p$. The parameter gradient in Equation~\eqref{eq:A5} and the logit gradient here express the same chain-rule differentiation in two coordinate systems. All coefficients are evaluated at the fixed rollout point.

This expression can also be verified using Equation~\eqref{eq:A9} and the centering identity:

\[
\begin{aligned}
\sum_b p_t(b)C_t(b)\nabla_{\ell_t}\log p_t(b)
&=\sum_b p_t(b)C_t(b)
\left[\operatorname{onehot}(b)-p_t\right]\\
&=p_t\odot C_t
-p_t\sum_b p_t(b)C_t(b)\\
&=p_t\odot C_t.
\end{aligned}
\]

Substitute the sum in Equation~\eqref{eq:A12} into $d_t=\langle g_t,k_t\rangle$ and use the inner product's linearity under addition and scalar multiplication:

\begin{equation}
\begin{aligned}
d_t
&=\left\langle
\sum_b p_t(b)\frac{C_t(b)}{\sigma}
\nabla_{\ell_t}\log p_t(b),\,k_t
\right\rangle\\
&=\sum_b p_t(b)\frac{C_t(b)}{\sigma}
\left\langle
\nabla_{\ell_t}\log p_t(b),\,
\nabla_{\ell_t}\log p_t(a_t)
\right\rangle.
\end{aligned}
\tag{A13}
\label{eq:A13}
\end{equation}

Equation~\eqref{eq:A13} explicitly connects the local reward advantages in the end-to-end gradient to directional credit. The inner product accounts for both the angle and magnitudes of the candidate-action and sampled-action directions. DATC then optimizes the weights using $A d_t$, yielding the final training coefficient $A w_t$. These are the two steps that convert a reward surrogate signal into a GRPO allocation.

Combining Equations~\eqref{eq:A6} and~\eqref{eq:A9} gives

\begin{equation}
\begin{aligned}
d_t
&=\frac1\sigma\sum_b p_t(b)C_t(b)
\left[\mathbf1\{b=a_t\}-p_t(b)\right]\\
&=\frac1\sigma\left[
p_t(a_t)C_t(a_t)-\sum_b p_t(b)^2C_t(b)
\right].
\end{aligned}
\tag{A14}
\label{eq:A14}
\end{equation}

The squared term arises because both the reward gradient and the sampled-action direction contain probability components. The centering identity cancels $\sum_b p_t(b)C_t(b)$, so the squared-probability-weighted term over the full vocabulary remains.

The evaluation above treats the logits at each position as independent coordinates. The actual actor changes these coordinates through shared parameters. Let $L_t=\partial\ell_t/\partial\theta$. An ordinary policy gradient ascent step is

\[
\Delta\theta
=\frac{\eta A}{T}\sum_u w_u L_u^\top k_u.
\]

It induces the logit change

\begin{equation}
\Delta\ell_t
=\frac{\eta A}{T}\sum_u w_u L_tL_u^\top k_u
+O(\eta^2),
\tag{A15}
\label{eq:A15}
\end{equation}

with the corresponding first-order prediction of surrogate reward

\begin{equation}
\widehat{\Delta(R/\sigma)}_{\mathrm{param}}
=\frac{\eta A}{T}\sum_u w_u
\sum_t g_t^\top L_tL_u^\top k_u.
\tag{A16}
\label{eq:A16}
\end{equation}

Shared parameters couple different positions through $L_tL_u^\top$. DATC uses the local logit utility in Equation~\eqref{eq:A11} to construct token-level credit, and the local gain result in the main text pertains to this evaluation model.

\subsection{KL-Based Weight Allocation and Local Gains}
\label{appendix:theory:kl-allocation}

The signed utility $A d_t$ measures the predicted gain at each position, while KL regularization controls how far the learning shares deviate from original GRPO. Together, they determine a unique softmax allocation whose additional predicted gain over the original weights at least covers its KL cost. For a fixed response's $A,d$ and a finite temperature $\tau>0$, define

\begin{equation}
F_\tau(w)=\frac AT\sum_t w_td_t
-\frac\tau T\sum_t w_t\log w_t,
\qquad
w_t\ge0,\quad\sum_t w_t=T.
\tag{A17}
\label{eq:A17}
\end{equation}

We use the continuous extension $0\log0=0$. The constraints make $w/T$ a distribution over positions, and the regularizer satisfies

\[
\frac1T\sum_t w_t\log w_t
=D_{\mathrm{KL}}\left(w/T\,\|\,\mathbf1/T\right).
\]

Introducing a Lagrange multiplier $\nu$, an interior stationary point satisfies

\[
\frac{\partial}{\partial w_t}
\left[
F_\tau(w)+\nu\left(\sum_jw_j-T\right)
\right]
=\frac1T\left[A d_t-\tau(\log w_t+1)\right]+\nu=0.
\]

Using $\sum_t w_t=T$ gives

\begin{equation}
w_t^*=T\frac{\exp(A d_t/\tau)}{Z},
\qquad
Z=\sum_j\exp(A d_j/\tau).
\tag{A18}
\label{eq:A18}
\end{equation}

To also account for feasible weights on the boundary, substitute $\log w_t^*=\log T+A d_t/\tau-\log Z$, obtaining

\begin{equation}
F_\tau(w)
=\tau\log(Z/T)
-\tau D_{\mathrm{KL}}\left(w/T\,\|\,w^*/T\right).
\tag{A19}
\label{eq:A19}
\end{equation}

Therefore, $F_\tau(w^*)-F_\tau(w)\ge0$, with equality if and only if $w=w^*$. Equation~\eqref{eq:A18} is the unique global optimum.

Finite utilities and a positive temperature imply $w_t^*>0$. Consequently,

\begin{equation}
\operatorname{sign}(\widetilde A_t)=\operatorname{sign}(A),
\qquad
\frac1T\sum_t\widetilde A_t
=\frac AT\sum_t w_t^*=A.
\tag{A20}
\label{eq:A20}
\end{equation}

These two properties ensure that DATC preserves the sign and mean of the response-level advantage when redistributing training coefficients within a response.

The original weights $\mathbf1$ are feasible in Equation~\eqref{eq:A17}, and their allocation KL is zero. Since $F_\tau(w^*)\ge F_\tau(\mathbf1)$,

\begin{equation}
\frac AT\sum_t(w_t^*-1)d_t
\ge\frac\tau T\sum_t w_t^*\log w_t^*
=\tau D_{\mathrm{KL}}\left(w^*/T\,\|\,\mathbf1/T\right)
\ge0.
\tag{A21}
\label{eq:A21}
\end{equation}

Multiplying by the step size $\eta>0$ and using Equation~\eqref{eq:A11} yields Equation~\eqref{eq:10} in the main text. When $A d_t$ varies across positions, the optimal weights are nonuniform and the difference in predicted gains is strictly positive. When the utilities are constant, the optimal weights are $\mathbf1$. In particular, when $A=0$, all final token-level advantages are zero.

\subsection{Adaptive Temperature and Weight Range}
\label{appendix:theory:adaptive-temperature}

The main text uses $\lambda$ to specify the maximum amplification and then selects the corresponding temperature. This section proves that such a choice is feasible and describes how to find it.

For response $i$, define

\begin{equation}
w_{it}^*(\tau)
=T_i\frac{\exp(A_i d_{it}/\tau)}
{\sum_j\exp(A_i d_{ij}/\tau)},
\qquad
M_i(\tau)=\max_t w_{it}^*(\tau).
\tag{A22}
\label{eq:A22}
\end{equation}

The positions with the highest utility have the largest weight at every positive temperature. Let $m_i=\max_t A_i d_{it}$. Differentiating at any highest-utility position gives

\begin{equation}
\frac{dM_i}{d\tau}
=-\frac{M_i(\tau)}{\tau^2}
\left[
m_i-\sum_t\frac{w_{it}^*(\tau)}{T_i}A_i d_{it}
\right]\le0.
\tag{A23}
\label{eq:A23}
\end{equation}

When utilities are nonconstant, the bracketed term is strictly positive, so the maximum weight strictly decreases. Tied highest-utility positions have identical weight functions, so the expression still holds. When utilities are constant or $T_i=1$, all weights equal $1$. Moreover,

\[
\lim_{\tau\to\infty}M_i(\tau)=1.
\]

Thus, for any $\lambda>1$, a feasible temperature exists for Equation~\eqref{eq:11}. If $M_i(1)\le\lambda$, the smallest feasible temperature is $1$. If $M_i(1)>\lambda$, the equation $M_i(\tau_i)=\lambda$ has a unique solution with $\tau_i>1$.

Let $\Delta u_i=\max_t A_i d_{it}-\min_t A_i d_{it}$. Lower-bounding each term in the denominator gives

\[
M_i(\tau)
=\frac{T_i\exp(m_i/\tau)}
{\sum_t\exp(A_i d_{it}/\tau)}
\le\exp(\Delta u_i/\tau).
\]

Hence,

\begin{equation}
\tau_{i,\mathrm{hi}}
=\max\left\{1,\frac{\Delta u_i}{\log\lambda}\right\}
\tag{A24}
\label{eq:A24}
\end{equation}

is a feasible upper bound.

Starting from temperature $1$ and repeatedly doubling it while the constraint remains unsatisfied finds a feasible upper endpoint and an adjacent infeasible lower endpoint. We then take the arithmetic midpoint of the two endpoints, update the interval according to whether the maximum-weight constraint is satisfied, and ultimately return the feasible endpoint. The implementation uses 40 bisection iterations and processes responses in batches. Each iteration halves the temperature interval's width, with the final precision determined by floating-point representation. Equation~\eqref{eq:A24} establishes the existence of a finite feasible upper bound; the implementation finds it through doubling.

The experiments additionally require a minimum weight of at least $1/\lambda$. This extension uses the same doubling and bisection procedure, checking both bounds when testing feasibility. Its monotonicity is proved in Appendix~\ref{appendix:theory:practical-allocation}.

Adaptive temperature converts $\lambda$ into a per-response KL regularization strength. For a selected and cached $\tau_i$, Equation~\eqref{eq:A18} remains the unique optimum of $F_{\tau_i}$, and the local gain comparison in Appendix~\ref{appendix:theory:kl-allocation} continues to hold.

\subsection{Correspondence between the Experimental Allocator and the Theory}
\label{appendix:theory:practical-allocation}

Before allocation, the experiments standardize directional credit within each response and fix the weights of a small number of formatting positions at $1$. Both operations preserve the KL-optimal allocation structure above: centering cancels in the softmax, scale standardization changes the effective temperature, and fixed positions restrict the set of positions available for allocation.

Let $\mathcal F$ be the set of free positions participating in allocation, with $N=|\mathcal F|>0$. Within-response standardization can be written as

\begin{equation}
\widehat d_t=\frac{d_t-\bar d}{\kappa},
\qquad
\bar d=\frac1N\sum_{t\in\mathcal F}d_t,
\qquad \kappa>0.
\tag{A25}
\label{eq:A25}
\end{equation}

Here $\kappa$ is a response-level scale held fixed during credit computation; its numerically stable form is given in Appendix~\ref{appendix:training:allocator}. Since $A\bar d/\kappa$ is shared by all free positions,

\begin{equation}
\begin{aligned}
w_t
&=N\operatorname{softmax}_{t\in\mathcal F}
\left(\frac{A\widehat d}{\tau}\right)\\
&=N\operatorname{softmax}_{t\in\mathcal F}
\left(\frac{A d}{\tau_{\mathrm{eff}}}\right),
\qquad \tau_{\mathrm{eff}}=\kappa\tau.
\end{aligned}
\tag{A26}
\label{eq:A26}
\end{equation}

The remaining valid response positions use $w_t=1$, so the full response still satisfies $\sum_t w_t=T$. Substituting the fixed positions into Equation~\eqref{eq:A17}, they contribute only constants independent of the free weights. Under $\sum_{t\in\mathcal F}w_t=N$, Equation~\eqref{eq:A26} is therefore the unique optimum at temperature $\tau_{\mathrm{eff}}$. Thus, the base temperature of $1$ after standardization corresponds to temperature $\kappa$ in the original credit coordinates.

The experiments increase $\tau$ until free positions simultaneously satisfy $1/\lambda\le w_t\le\lambda$. Write $u_t=A\widehat d_t$ and $q_t=w_t/N$. The lowest-utility position has the minimum weight $m(\tau)$, whose derivative is

\begin{equation}
\frac{dm}{d\tau}
=-\frac{m(\tau)}{\tau^2}
\left[\min_t u_t-\sum_{t\in\mathcal F}q_tu_t\right]
\ge0.
\tag{A27}
\label{eq:A27}
\end{equation}

Together with Equation~\eqref{eq:A23}, this shows that as temperature increases, the minimum weight increases and the maximum weight decreases, both approaching $1$. A finite feasible temperature therefore also exists for the two-sided constraints. Once the temperature is selected, Equation~\eqref{eq:A26} remains the optimum of the corresponding KL objective; the two-sided range is enforced by choosing the regularization strength. Using $\mathbf1$ as a feasible reference gives

\begin{equation}
\widehat{\Delta(R/\sigma)}(w)
-\widehat{\Delta(R/\sigma)}(\mathbf1)
\ge
\frac{\eta\tau_{\mathrm{eff}}}{T}
\sum_{t\in\mathcal F}w_t\log w_t
\ge0.
\tag{A28}
\label{eq:A28}
\end{equation}

This result has the same local meaning as in the main text: it compares first-order reward increments at a fixed response in independent logit coordinates. The correspondence to shared-parameter updates is given by Equation~\eqref{eq:A16}.

When the free-position set is empty or the utilities are constant, the allocator returns the original weights $\mathbf1$.

\clearpage
\section{Training Objective and Implementation}
\label{appendix:training:overview}

DATC computation has two stages: first compute and cache token-level advantages at sampled responses, then update the policy using these coefficients. This organization separates RM input backpropagation from actor parameter updates while retaining the directional credit derived from the end-to-end gradient in the main text. This appendix provides the complete objective and allocation procedure used in the experiments. Hyperparameters are listed in Appendix~\ref{appendix:experiments:setup}.

\subsection{From Response Rewards to the Policy Objective}
\label{appendix:training:objective}

For $G$ responses to the same input, we first compute the raw scores $R_i$ from the frozen RM. All compared methods use the same length-dependent reward processing, imposing a moderate cost on very short and excessively long responses. Let $L_i$ be the number of generated tokens and $\sigma_0$ a reward scale fixed in advance for the model setting. Then

\begin{equation}
\begin{aligned}
P_{\mathrm{short}}(L_i)
&=0.5\sigma_0\left[\frac{8-L_i}{8}\right]_0^1,\\
P_{\mathrm{long}}(L_i)
&=2\sigma_0\left[\frac{L_i-1024}{2048-1024}\right]_0^1,\\
r_i&=R_i-P_{\mathrm{short}}(L_i)-P_{\mathrm{long}}(L_i).
\end{aligned}
\tag{B1}
\label{eq:B1}
\end{equation}

Here $[z]_0^1=\min(1,\max(0,z))$. Responses of $8$ to $1024$ tokens retain their raw scores. Length counts include the generated termination token but exclude prompt and padding tokens. Empty responses and degenerate responses containing at least $32$ consecutive newlines receive the minimum reward among valid responses in the group minus $1$. If the entire group is degenerate, the input group is resampled. We then compute

\begin{equation}
\bar r=\frac1G\sum_i r_i,
\qquad
\sigma=\max\left\{
\sqrt{\frac1G\sum_i(r_i-\bar r)^2},\;0.5\sigma_0
\right\},
\qquad
A_i=\frac{r_i-\bar r}{\sigma}.
\tag{B2}
\label{eq:B2}
\end{equation}

The lower bound on the scale controls advantages when rewards are very similar. The four model settings use $\sigma_0=3.0323$ for Qwen Base, $1.2118$ for Qwen Instruct, $6.2743$ for Llama Base, and $5.9262$ for Llama Instruct. All methods within a setting use the same value. Directional credit is still computed from the input gradient of the raw RM score and divided by the group scale in Equation~\eqref{eq:B2}. With the sampled response and its length fixed, the length terms in Equation~\eqref{eq:B1} are constant, so their input derivatives in the local expansion are zero.

Write $\widetilde A_{it}=A_iw_{it}$. Let $\pi_{\mathrm{old}}$ be the policy before the update and $\pi_{\mathrm{ref}}$ the reference policy frozen at the start of reinforcement learning. The reference policy in the Base settings includes the lightweight SFT adapter; the Instruct settings use the corresponding initial instruction-tuned model. The policy loss is

\begin{equation}
\begin{aligned}
\mathcal L_{\mathrm{policy}}
&=-\frac1B\sum_{i=1}^B\frac1{T_i}\sum_{t=1}^{T_i}
c_{it}\min\left\{
\rho_{it}\widetilde A_{it},
\operatorname{clip}(\rho_{it},1-\epsilon,1+\epsilon)
\widetilde A_{it}
\right\},\\
\rho_{it}
&=\frac{\pi_\theta(a_{it}\mid s_{it})}
{\pi_{\mathrm{old}}(a_{it}\mid s_{it})}.
\end{aligned}
\tag{B3}
\label{eq:B3}
\end{equation}

The experiments generate responses with vLLM and recompute old-policy probabilities on the training backend. To account for token-probability differences between the two execution backends, we use the cached coefficient

\begin{equation}
c_{it}=\operatorname{sg}\left[
\frac{\pi_{\mathrm{old}}(a_{it}\mid s_{it})}
{\pi_{\mathrm{sample}}(a_{it}\mid s_{it})}
\right].
\tag{B4}
\label{eq:B4}
\end{equation}

This coefficient multiplies the clipped policy term from outside the clipping operation and is also used in the reference-policy KL term. It corrects token probabilities conditional on the sampled prefix. The local direction derivation in the main text corresponds to the policy term when sampling probabilities and old-policy probabilities agree, in which case $c_{it}=1$. The full loss uses a common nonnegative KL estimator:

\begin{equation}
\begin{aligned}
\delta_{it}
&=\log\pi_{\mathrm{ref}}(a_{it}\mid s_{it})
-\log\pi_\theta(a_{it}\mid s_{it}),\\
\mathcal L(\theta)
&=\mathcal L_{\mathrm{policy}}
+\frac\beta B\sum_i\frac1{T_i}\sum_t
c_{it}\left[\exp(\delta_{it})-1-\delta_{it}\right].
\end{aligned}
\tag{B5}
\label{eq:B5}
\end{equation}

Both terms are first averaged over valid positions within each response and then over responses. During the policy update, $A_i$, $w_{it}$, $c_{it}$, old-policy probabilities, and reference probabilities are all fixed. GRPO uses the same objective with $w_{it}=1$, while DATC replaces these weights with those derived from directional credit.

\subsection{Input Gradients, Vocabulary Reductions, and Position Correspondence}
\label{appendix:training:credit-computation}

A single RM input backward pass provides $f_t=\nabla_{e_t}R$ at every position in the response. RM parameters remain frozen, while input embeddings retain gradients. The implementation obtains $f_t$ and old-policy probabilities separately and then computes the matrix products in the chain rule. This allows the RM computation graph to be released before the actor update.

Although the full vocabulary is typically large, Equation~\eqref{eq:7} requires only a few reduced quantities. Let $v_t(b)=\langle W_b,f_t\rangle$ and $\mu_t=\sum_b p_t(b)v_t(b)$. Then

\begin{equation}
d_t=\frac1\sigma\left[
p_t(a_t)\bigl(v_t(a_t)-\mu_t\bigr)
-\sum_b p_t(b)^2v_t(b)
+\mu_t\sum_b p_t(b)^2
\right].
\tag{B6}
\label{eq:B6}
\end{equation}

The implementation computes $v_t$ in blocks, accumulates $\mu_t$, $\sum_bp_t(b)^2v_t(b)$, and $\sum_bp_t(b)^2$, and extracts the component corresponding to the sampled token. The blocks cover the full candidate vocabulary, yielding the same reduction as an unblocked computation. Each block processes $128$ response positions and $8192$ vocabulary candidates in the experiments. Probabilities, gradient projections, and these reductions use FP32.

This step requires consistent candidate-token identities between the actor and RM. We check the token-to-ID mappings and special tokens of their tokenizers and map both vocabularies to shared token identities. The actor uses its own conversational input, while the RM receives the same prompt and response in its scoring template. Correspondence between response-text positions is used to extract the RM input gradient for each sampled token. Thus, tokenizer alignment ensures consistency in candidate-token semantics, while response-position mapping accounts for position offsets introduced by the two conversational templates.

Termination tokens, whitespace-only tokens, tokens containing newlines that reduce to periods after whitespace removal, and valid tokens without reliable position mappings receive a fixed weight of $1$. Other mapped positions participate in credit allocation. Ordinary words and general punctuation are treated as content positions. Response formatting therefore continues to learn with the original GRPO coefficient, while the corresponding update budget is distributed among content positions.

\subsection{Within-Response Standardization and Adaptive Temperature}
\label{appendix:training:allocator}

Gradient scales vary across responses, so directly using a fixed temperature would make weight concentration fluctuate accordingly. The experiments first standardize directional credit over the free-position set $\mathcal F$ and then use $\lambda$ to control the allocation range. Let $N=|\mathcal F|$. The implementation uses the following numerically stable form:

\begin{equation}
\begin{aligned}
m&=\max\left\{\max_{t\in\mathcal F}|d_t|,\varepsilon_{\mathrm{fp32}}\right\},
\qquad z_t=d_t/m,\\
\bar z&=\frac1N\sum_{t\in\mathcal F}z_t,
\qquad s_z=\sqrt{\frac1N\sum_{t\in\mathcal F}(z_t-\bar z)^2},\\
\widehat d_t
&=\frac{z_t-\bar z}
{s_z+10^{-3}\frac1N\sum_{t\in\mathcal F}|z_t|+\varepsilon_{\mathrm{fp32}}},
\qquad u_t=A\widehat d_t.
\end{aligned}
\tag{B7}
\label{eq:B7}
\end{equation}

Here $\varepsilon_{\mathrm{fp32}}$ is the smallest positive normal FP32 number. Dividing by $m$ first avoids overflow when squaring. Equation~\eqref{eq:B7} is equivalent to $\widehat d_t=(d_t-\bar d)/\kappa$, where

\begin{equation}
\kappa=\sqrt{\frac1N\sum_{t\in\mathcal F}(d_t-\bar d)^2}
+10^{-3}\frac1N\sum_{t\in\mathcal F}|d_t|
+m\varepsilon_{\mathrm{fp32}}.
\tag{B8}
\label{eq:B8}
\end{equation}

The weights at free positions are

\begin{equation}
w_t(\tau)=N\operatorname{softmax}_{t\in\mathcal F}(u/\tau),
\qquad
\tau_i=\min\left\{\tau\ge1:
\min_{t\in\mathcal F}w_t(\tau)\ge\frac1\lambda,
\ \max_{t\in\mathcal F}w_t(\tau)\le\lambda
\right\}.
\tag{B9}
\label{eq:B9}
\end{equation}

The experiments use $\lambda$ to control the maximum amplification and its reciprocal to control the minimum weight. With $\lambda=4$ in the main experiments, weights at positions participating in allocation lie in $[0.25,4]$, while the mean weight over the entire response remains $1$. Appendix~\ref{appendix:theory:practical-allocation} proves that within-response standardization can be absorbed into the effective temperature and that the two-sided constraints preserve temperature monotonicity and the corresponding KL optimum.

The search starts at temperature $1$. Responses already satisfying the range retain that temperature. For the others, we repeatedly double the upper endpoint and, after finding a feasible interval, perform $40$ ordinary bisection iterations, returning the feasible endpoint. Weights are set to $1$ when there are no free positions, only one free position, or constant utilities. Finally, $\widetilde A_t=Aw_t$ is cached for the subsequent policy-update epoch.

\subsection{One Training Iteration}
\label{appendix:training:algorithm}

Algorithm~\ref{algorithm:appendix-training:rollout} summarizes one DATC rollout. Credit computation produces fixed token-level advantages, which are then used to train the actor's LoRA parameters during the policy update.

\begin{algorithm}[tb]
\caption{One training iteration of DATC.}
\label{algorithm:appendix-training:rollout}
\begin{algorithmic}[1]
\Require Actor $\pi_\theta$ and its synchronized sampler $\pi_{\mathrm{sample}}$; frozen RM $R_\phi$ and reference policy $\pi_{\mathrm{ref}}$; a batch of prompts, group size $G$, and weight upper bound $\lambda>1$.
\Ensure Updated actor and synchronized sampler.
\State Save the old policy $\pi_{\mathrm{old}}\gets\pi_\theta$.
\For{each prompt $x$ in the batch}
    \State Sample $G$ responses; save tokens, masks, and sampling probabilities.
    \State Score with $R_\phi$ and process responses as in Appendix~\ref{appendix:training:objective}.
    \State Compute adjusted rewards $r_i$, group scale $\sigma$, and advantages $A_i$ using Equations~\eqref{eq:B1}--\eqref{eq:B2}.
    \For{each response $y_i=(a_{i1},\ldots,a_{iT_i})$}
        \State Recompute $p_{it}=\pi_{\mathrm{old}}(\cdot\mid s_{it})$ on the training backend.
        \State Compute $\pi_{\mathrm{ref}}(a_{it}\mid s_{it})$ and cache correction coefficients $c_{it}$ using Equation~\eqref{eq:B4}.
        \State Set $w_{it}\gets1$ at all valid response positions.
        \State Determine the free-position set $\mathcal F_i$ as in Appendix~\ref{appendix:training:credit-computation}.
        \If{$\lvert\mathcal F_i\rvert>1$}
            \State Compute $f_{it}=\nabla_{e_{it}}R_\phi(x,y_i)$ with one RM input backward pass.
            \State Map input gradients to response positions; compute $d_{it}$ using Equation~\eqref{eq:B6}.
            \State Stop gradients through $d_{it}$ and release the RM computation graph.
            \State Standardize $d_{it}$ over $\mathcal F_i$ using Equation~\eqref{eq:B7}; set $u_{it}\gets A_i\widehat d_{it}$.
            \If{$u_{it}$ is nonconstant over $\mathcal F_i$}
                \State Find the smallest feasible $\tau_i\ge1$ by bisection as in Appendix~\ref{appendix:training:allocator}.
                \State Set $w_{it}\gets\lvert\mathcal F_i\rvert\operatorname{softmax}_{t\in\mathcal F_i}(u_i/\tau_i)$ for $t\in\mathcal F_i$.
            \EndIf
        \EndIf
        \State Set $\widetilde A_{it}\gets\operatorname{sg}(A_iw_{it})$.
        \State Cache $\widetilde A_{it}$, $c_{it}$, and old-policy and reference-policy probabilities of sampled tokens.
    \EndFor
\EndFor
\For{each minibatch in one policy-update epoch}
    \State Construct $\mathcal L(\theta)$ using cached coefficients and Equations~\eqref{eq:B3}--\eqref{eq:B5}.
    \State $\theta\gets\operatorname{AdamWStep}(\theta,\nabla_\theta\mathcal L)$.
\EndFor
\State Synchronize the updated LoRA adapter to the sampler.
\end{algorithmic}
\end{algorithm}

In the experiments, the actor, RM, and sampler each occupy one GPU. The actor and RM backbones use BF16, while the actor output head and probabilities used for credit computation use FP32. Gradient checkpointing is enabled for the actor. The additional computation is concentrated in RM input backpropagation and vocabulary reductions; end-to-end training overhead is reported in Appendix~\ref{appendix:experiments:efficiency}.

\clearpage
\section{Experimental Setup and Supplementary Analyses}
\label{appendix:experiments:overview}

This appendix provides training configurations and evaluation details, a human evaluation protocol, and supplementary analyses of training dynamics, computational costs, and shared-parameter single-step gains.

\subsection{Experimental Setup and Comparison Implementations}
\label{appendix:experiments:setup}

We construct the experimental data from \texttt{HuggingFaceH4/ultrafeedback\_binarized}. Text normalization, deduplication, and length filtering yield approximately 59,000 candidate training examples. SFT uses a fixed set of 2,500 instruction--response examples, and RL uses 2,000 prompts. RM-Reward evaluation uses 256 fixed prompts from the held-out set. SFT uses the dataset's chosen responses, while RL generates responses online from the current policy.

We select Qwen3 and Llama 3.1 to cover different model families and pretraining and post-training backgrounds. For each family, we evaluate both the Base model after lightweight SFT and the official Instruct model. The former examines preference learning after basic instruction adaptation; the latter examines further optimization on top of existing alignment capabilities. Table~\ref{tab:checkpoints} lists the checkpoints. Qwen3-14B-Instruct in the main text refers to the official \texttt{Qwen3-14B} instruction-tuned model. The two model families use Skywork-Reward-V2-Qwen3-8B~\citep{liu2026skywork} and Skywork-Reward-Llama-3.1-8B-v0.2~\citep{liu2024skywork}, respectively. DATC requires tokenizer alignment between the actor and RM so that candidate-token probabilities and their corresponding rows in the RM embedding matrix have the same meaning. The implementation further aligns token IDs and response positions.

\begin{table}[hbp]
\caption{Policy models, reward models, and initializations.}
\label{tab:checkpoints}
\centering
\begin{tabularx}{\linewidth}{@{}>{\raggedright\arraybackslash}p{0.14\linewidth}>{\raggedright\arraybackslash}p{0.23\linewidth}>{\raggedright\arraybackslash}X>{\raggedright\arraybackslash}p{0.17\linewidth}@{}}
\toprule
Setting & Policy checkpoint & Reward model checkpoint & RL initialization \\
\midrule
Qwen Base & Qwen3-14B-Base & Skywork-Reward-V2-Qwen3-8B & After lightweight SFT \\
Qwen Instruct & Qwen3-14B & Skywork-Reward-V2-Qwen3-8B & Official instruction-tuned model \\
Llama Base & Llama-3.1-8B & Skywork-Reward-Llama-3.1-8B-v0.2 & After lightweight SFT \\
Llama Instruct & Llama-3.1-8B-Instruct & Skywork-Reward-Llama-3.1-8B-v0.2 & Official instruction-tuned model \\
\bottomrule
\end{tabularx}
\end{table}

Both Base models are trained on the SFT data above for 2 epochs, with a learning rate of $10^{-4}$, a maximum sequence length of 4,096, a microbatch size of 4, 8 gradient accumulation steps, and a warmup ratio of 0.03. GRPO and DATC for the same model start from the same SFT checkpoint. Instruct settings proceed directly to reinforcement learning. RL processes the selected 2,000 prompts in a fixed order, 8 at a time, for 250 rollouts. Each prompt has 8 sampled responses, and each rollout is followed by one policy update. Evaluations in the main text use the checkpoint at the end of training.

The final evaluations measure reward under the training RM, preferences over open-ended responses, and multi-turn response quality. RM-Reward uses 256 independent held-out prompts, with generated responses scored by the corresponding RM, and reports the mean before length penalties. Because score scales differ across RMs, this metric is interpreted through paired comparisons within a model family. AlpacaEval uses the dataset-provided GPT-4 Turbo~\citep{hurst2024gpt} (\texttt{gpt4\_1106\_preview}) responses as references and GPT-4.1~\citep{hurst2024gpt} as the judge, reporting the raw win rate as the mean judge-assigned preference probability, following the official AlpacaEval aggregation procedure. Arena-Hard v2.0 uses the official \texttt{gpt-4o-mini-2024-07-18}~\citep{achiam2023gpt} responses as references and GPT-4.1 as the judge. Each pair is judged twice with the response order swapped, and the reported win rate is weighted by outcome strength: wins, ties, and losses receive scores of 1, 0.5, and 0, respectively, with a weight of 3 for decisive wins or losses and 1 otherwise. MT-Bench uses the official 80 two-turn conversations. GPT-4o scores each turn on a 1--10 scale, and the final score is the equally weighted mean of the 160 ratings. Mathematics, reasoning, and coding questions use the official GPT-4 reference responses. In the main text, AlpacaEval and Arena-Hard win rates are reported as percentages, and differences between methods are expressed in percentage points. RM-Reward and MT-Bench retain their respective score scales. The main experiments compare DATC with GRPO using the same training initialization, RM, and policy optimization settings. Supplementary comparisons on Qwen Base further distinguish the roles of credit sources, directional relationships, and position correspondence. RED provides a reward-redistribution baseline based on prefix-score differences, while P2T provides a token reward attribution baseline based on RM input gradients. Together, these comparisons evaluate DATC against different approaches to RM-based token credit assignment.

DATC results in Tables~\ref{tab:main-results} and~\ref{tab:credit-ablation} are reported as means over three random seeds, with standard deviations additionally reported in Table~\ref{tab:main-results}.

\begin{table}[tb]
\caption{Shared training settings.}
\label{tab:training-settings}
\centering
\begin{tabularx}{\linewidth}{@{}XX@{}}
\toprule
Configuration & Value \\
\midrule
Prompts per rollout / responses per prompt & 8 / 8 \\
Rollouts / policy epochs per rollout & 250 / 1 \\
Maximum prompt / response length & 2,048 / 2,048 tokens \\
Sampling temperature / top-$p$ / top-$k$ & 1.0 / 1.0 / 0 (full distribution) \\
Optimizer / learning rate / weight decay & AdamW / constant $5\times10^{-5}$ / 0.01 \\
Maximum gradient norm / clipping parameter $\epsilon$ & 1.0 / 0.2 \\
KL coefficient $\beta$ / reference policy & 0.03 / frozen initial RL policy \\
Update minibatch / microbatch & 64 responses / 1 response \\
RM gradient microbatch & 1 response \\
LoRA rank / alpha / dropout & 64 / 128 / 0 \\
LoRA modules & q, k, v, o, gate, up, down projections \\
Backbone precision / policy output-head and probability precision & BF16 / FP32 \\
Base allocation temperature & 1 \\
Main-experiment $\lambda$ / ablation values & 4 / 2, 4, 8 \\
\bottomrule
\end{tabularx}
\end{table}

The main experiments use 3 H200 GPUs. Policy updates use LoRA, freezing the original model weights and training low-rank adaptation parameters. Gradient checkpointing is enabled for the actor. vLLM uses single-GPU tensor parallelism, a maximum context length of 4,096, a maximum of 32 concurrent sequences, and a GPU memory utilization parameter of 0.45. The training software is PyTorch 2.13.0+cu129 and Transformers 5.16.1. All methods use the same length-dependent rewards and degenerate-response processing.

The direction-alignment ablation removes the angle term from $d_t=\|g_t\|_2\|k_t\|_2\cos\alpha_t$, replacing the directional credit passed to the allocator with $d_t^{\mathrm{norm}}=\|g_t\|_2\|k_t\|_2.$

After replacement, the variant retains the within-response standardization over free positions in Appendix~\ref{appendix:training:allocator}. The standardized signal is multiplied by the response-level advantage $A$ to form the allocation utility. It uses the same adaptive-temperature search and two-sided weight constraints, with $\lambda=4$. Fixed positions specified in Appendix~\ref{appendix:training:credit-computation} retain weight $1$. This variant changes only the credit signal and preserves the subsequent allocation procedure, testing the role of directional relationships relative to gradient magnitudes. The shuffled-weight variant permutes the resulting weights within each response. For a permutation $\rho$ over the corresponding position set, $w_t^{\mathrm{shuffle}}=w_{\rho(t)}$ preserves the response's weight set, mean, and extrema while changing the update share at each position. Its difference from full DATC directly tests the correspondence between weights and text positions. The weight-range experiment varies $\lambda$ over 2, 4, and 8 to examine how the allowed update concentration affects results.

\subsection{Reward and Response-Length Dynamics}
\label{appendix:experiments:dynamics}

\begin{figure}[htbp]
\centering
\includegraphics[width=\linewidth]{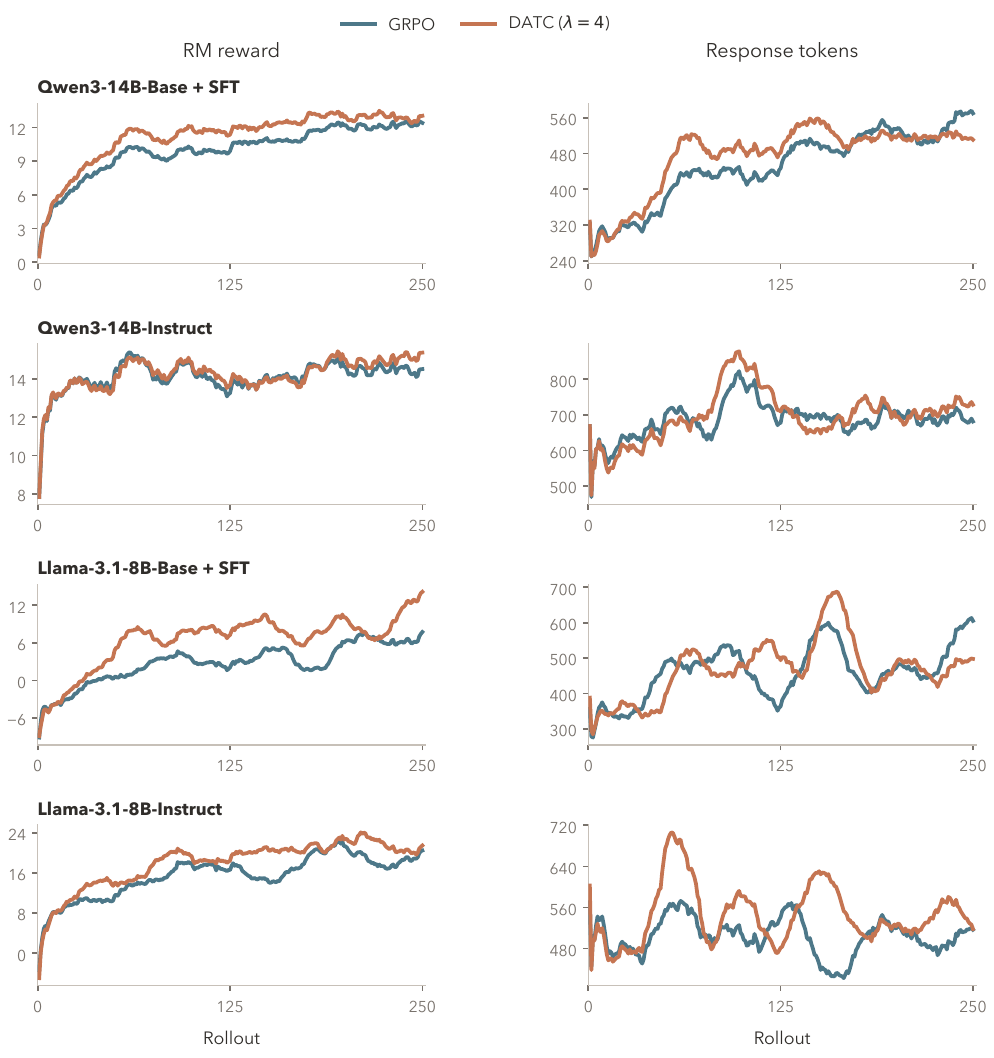}
\caption{Raw RM reward (left) and mean response length (right) of DATC and GRPO across the four experimental settings. Curves show trailing 25-rollout moving averages.}
\label{fig:training-all}
\end{figure}

Figure~\ref{fig:training-all} shows raw RM reward and response length in the Qwen Base, Qwen Instruct, Llama Base, and Llama Instruct settings. Within each setting, DATC is compared with GRPO using the same training initialization and rollout budget. Rewards in the training plots come from online sampled inputs, whereas RM-Reward in the main results table comes from fixed held-out prompts.

Over the final 25 rollouts, DATC achieves a higher mean raw reward than the corresponding GRPO baseline in all four settings. For Llama Base, the final-window mean rewards are 14.14 for DATC and 7.72 for GRPO, with mean response lengths of 497.9 and 605.2 tokens, respectively. DATC produces shorter responses in the final window for both Base settings. Response lengths are similar for Llama Instruct, while DATC produces longer responses for Qwen Instruct. These results show that DATC improves training reward in all four settings, with shorter responses in both Base settings as well.

\subsection{Training Overhead}
\label{appendix:experiments:efficiency}

\begin{figure}[htbp]
\centering
\includegraphics[width=\linewidth]{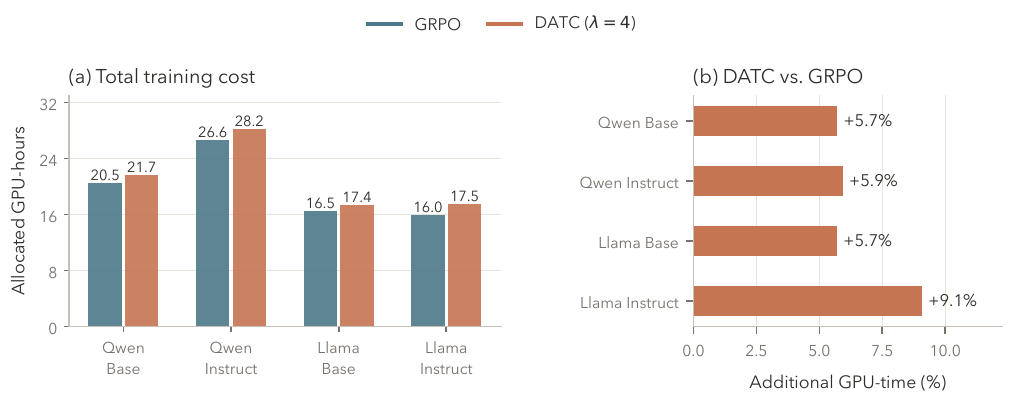}
\caption{Cumulative GPU time to complete training (left) and DATC's overhead relative to GRPO (right) across the four experimental settings. End-to-end costs include sampling, scoring, credit computation, and policy updates.}
\label{fig:compute-cost}
\end{figure}

We measure end-to-end cost using cumulative GPU time, defined as training wall-clock time multiplied by the number of allocated GPUs. It includes the total cost of sampling, scoring, credit computation, and policy updates under the same rollout budget. Figure~\ref{fig:compute-cost} summarizes total training cost across the four settings. DATC increases total GPU time by approximately 5.7\%--9.1\% over the corresponding GRPO baseline.

Together with Table~\ref{tab:main-results}, these results show that DATC achieves higher mean RM-Reward, AlpacaEval, Arena-Hard, and MT-Bench scores than the corresponding GRPO baseline in all four settings. For Qwen Base, a GPU-time increase of approximately 5.7\% accompanies AlpacaEval and Arena-Hard gains of 12.18 and 10.82 percentage points, respectively. DATC thus delivers performance improvements across model families and initialization settings with modest additional training cost.

\subsection{Human Evaluation}
\label{appendix:experiments:human}

\begin{wraptable}{r}{0.48\textwidth}
\centering
\caption{Human preference evaluation on Qwen3-14B-Base. Win rates are the percentages of 120 prompts won by each method after tie-breaking.}
\label{tab:human-evaluation}
\begin{tabularx}{\linewidth}{@{}Xc@{}}
\toprule
Method & Win rate (\%) $\uparrow$ \\
\midrule
DATC & \textbf{65.00} \\
GRPO & 16.67 \\
P2T & 15.00 \\
RED & 3.33 \\
\bottomrule
\end{tabularx}
\end{wraptable}

To complement automatic evaluations, we compare DATC ($\lambda=4$), GRPO, P2T, and RED on Qwen3-14B-Base using 120 held-out prompts excluded from training and hyperparameter selection. We randomly assign 40 prompts to each of the three training seeds. For each prompt, all methods use their final checkpoints from the assigned seed and identical decoding settings.

Seven evaluators proficient in English independently select the best of the four responses, presented under anonymous labels in randomized order, considering instruction following, correctness, relevance, completeness, and clarity. The method receiving the most votes wins that prompt. If multiple methods tie for the highest vote count, the same seven evaluators vote again among only the tied responses, with the winner determined by the most votes. Table~\ref{tab:human-evaluation} reports the percentage of prompts won by each method after tie-breaking.

In this four-way comparison, DATC wins 78 of the 120 prompts (65.00\%), compared with 20 (16.67\%) for GRPO, 18 (15.00\%) for P2T, and 4 (3.33\%) for RED. This result is consistent with DATC's higher AlpacaEval and Arena-Hard win rates in the same Qwen3-14B-Base setting. GRPO and P2T obtain similar win rates in human evaluation, differing by only two prompts, with both outperforming RED. These results provide complementary evidence that DATC's gains in automatic evaluations are accompanied by improvements in response quality as assessed by human evaluators.

\subsection{Independent-Logit Predictions and Shared-Parameter Single-Step Gains}
\label{appendix:experiments:single-step-gain}

The local gain analysis in the main text uses a surrogate model with independent logits, whereas actual policy updates change outputs at multiple positions through shared parameters. To examine the connection between these two descriptions, we compare DATC's independent-logit predicted gain over GRPO with the first-order surrogate gain associated with an actual single-step parameter update.

\textbf{Experimental setup.} We use a fixed Qwen3-14B-Base SFT checkpoint and a frozen Skywork-Reward-V2-Qwen3-8B, sampling 100 batches to obtain 100 data points. Each point corresponds to a batch of 8 prompts with 8 responses sampled per prompt. Every batch starts from the same model parameters and initial optimizer state. DATC and GRPO share the sampled responses, response-level advantages, and optimization settings, and each performs one AdamW update. We use LoRA updates with a learning rate of $5\times10^{-5}$ and a gradient norm limit of $1.0$. DATC uses $\lambda=4$.

\textbf{Gain measures.} For a batch of $B$ responses, the additional gain predicted by the independent-logit model is

\begin{equation}
X=\frac{1}{B}\sum_{i=1}^{B}\frac{A_i}{T_i}
\sum_{t=1}^{T_i}(w_{it}-1)d_{it},
\tag{C1}
\label{eq:C1}
\end{equation}

where $A_i$, $T_i$, $w_{it}$, and $d_{it}$ denote the response-level advantage, valid response length, DATC weight, and directional credit, respectively. GRPO assigns a weight of $1$. This quantity omits the common step-size factor.

Let $\theta_0$ be the shared starting point, and let $\theta_D$ and $\theta_G$ denote the parameters after the respective updates. With the sampled responses and their prefixes fixed, we compute the mean standardized reward gradient through the ST surrogate path defined in the main text and define the shared-parameter first-order gain difference as

\begin{equation}
h=\frac{1}{B}\sum_{i=1}^{B}
\frac{1}{\sigma_i}\widehat{\nabla}_{\theta}R_i(\theta_0),
\qquad
P=h^\top(\theta_D-\theta_G).
\tag{C2}
\label{eq:C2}
\end{equation}

Here, $\sigma_i$ is the reward normalization scale determined before the update and held fixed. Both the gradient and parameter difference are taken with respect to trainable LoRA parameters. The gradient is computed from the raw RM score, without differentiating the length penalty. A positive $P$ indicates that DATC's actual parameter update outperforms GRPO's under this first-order surrogate evaluation. Shared-parameter coupling, AdamW, and gradient clipping affect $P$. We use the correlation between $X$ and $P$ to assess how well the independent-logit prediction indicates shared-parameter update gains.

\begin{figure}[H]
\centering
\includegraphics[width=0.9\linewidth]{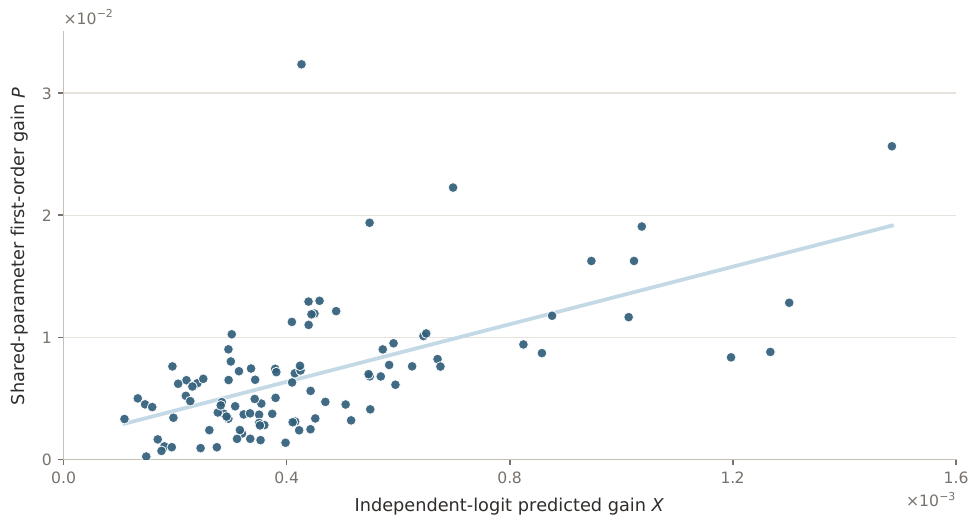}
\caption{Relationship between independent-logit predicted gain $X$ and shared-parameter single-step first-order gain $P$. Each point represents a within-batch DATC--GRPO comparison from the same SFT starting point, for a total of 100 data points. The light-colored solid line is a linear fit with an intercept.}
\label{fig:single-step-gain}
\end{figure}

\textbf{Results and discussion.} As shown in Figure~\ref{fig:single-step-gain}, across the 100 data points, $X$ and $P$ have a Spearman correlation of $0.619$ and a Pearson correlation of $0.589$, with $R^2=0.347$ for a linear fit with an intercept. All points satisfy $X>0$ and $P>0$. Thus, under the model initialization and single-step update setting studied here, DATC achieves higher shared-parameter first-order surrogate gains than GRPO in every batch. Larger independent-logit predicted gains also tend to correspond to larger shared-parameter first-order gains, supporting the use of directional credit to identify more beneficial update allocations in shared-parameter optimization.

\section{Limitations \& Future Work}
\label{appendix:limitations:overview}

DATC targets preference optimization with differentiable reward models. Computing directional credit requires access to RM input gradients and aligned vocabularies between the actor and RM. Future work could explore token- or sequence-alignment mechanisms across tokenizers, as well as gradient estimation for evaluators that provide only scores, to broaden the applicability of directional credit and evaluate how these approximations affect credit quality and computational cost.

Our experiments use LoRA to study general preference alignment under Base and Instruct initializations in two model families. Future evaluations could extend to full-parameter training, a wider range of model scales, and tasks with specialized reward models, such as mathematical reasoning and code generation, to examine the applicability of directional credit across training regimes and reward structures.

\end{document}